\documentclass[10pt,twocolumn,letterpaper]{article}

\usepackage{bm}
\newcommand{\realR}{\mathbb{R}}
\newcommand{\point}{\bm{p}}
\newcommand{\kernelPoint}{\bm{k}}
\newcommand{\weight}{\bm{w}}
\newcommand{\feature}{\bm{f}}
\newcommand{\neighbor}{\mathcal{N}}

\newcommand{\globalFeature}{\feature^{\text{global}}}

\newcommand{\outputFeature}{\feature^{\text{out}}}
\newcommand{\linearLayer}{\mathbf{h}}
\newcommand{\geometricLayer}{\bm{g}}
\newcommand{\pointCloud}{\mathcal{P}}

\newcommand{\rotation}{\mathbf{R}}
\newcommand{\translation}{\bm{t}}
\newcommand{\size}{\bm{s}}

\newcommand{\fiveDfiveC}{5^\circ5\text{cm}}

\newcommand{\fiveDtwoC}{5^\circ2\text{cm}}

\newcommand{\tenDfiveC}{10^\circ5\text{cm}}
\newcommand{\IoUSevenFive}{\text{IoU}_{75}}

\newcommand{\norm}[1]{\left\lVert#1\right\rVert}

\newcommand{\distance}{\text{dist}}

\makeatletter
\newcommand\footnoteref[1]{\protected@xdef\@thefnmark{\ref{#1}}\@footnotemark}
\makeatother

\usepackage[pagenumbers]{cvpr} 

\usepackage[accsupp]{axessibility}
\usepackage{graphicx}
\usepackage{amsmath}
\usepackage{amssymb}
\usepackage{booktabs}
\usepackage{tabularx}
\usepackage{multicol}
\usepackage{multirow}
\usepackage{tablefootnote}

\usepackage[pagebackref,breaklinks,colorlinks]{hyperref}

\usepackage[capitalize]{cleveref}
\crefname{section}{Sec.}{Secs.}
\Crefname{section}{Section}{Sections}
\Crefname{table}{Table}{Tables}
\crefname{table}{Tab.}{Tabs.}

\def\cvprPaperID{7985} 
\def\confName{CVPR}
\def\confYear{2023}

\begin{document}

\title{HS-Pose: Hybrid Scope Feature Extraction for Category-level Object Pose Estimation}

\author{
Linfang Zheng$^{1,4}$ \and Chen Wang$^{1,2}$ \and Yinghan Sun$^{1}$ \and Esha Dasgupta$^{4}$ \and Hua Chen$^{1}$ \and Ale\v{s} Leonardis$^{4}$ \and Wei Zhang\thanks{The corresponding author.} $^{1,3}$ \and Hyung Jin Chang$^{4}$ \and\\$^{1}$Department of Mechanical and Energy Engineering, Southern University of Science and Technology\\
$^{2}$Department of Computer Science, the University of Hong Kong\\
$^{3}$Peng Cheng Laboratory, Shenzhen, China\\
$^{4}$School of Computer Science, University of Birmingham\\
{\tt\small$\{$lxz948,exd949$\}$@student.bham.ac.uk, cwang5@cs.hku.hk, sunyh2021@mail.sustech.edu.cn}\\
{\tt\small $\{$chenh6,zhangw3$\}$@sustech.edu.cn,$\{$a.leonadis,h.j.chang$\}$@bham.ac.uk}
}


\maketitle

\begin{abstract}
In this paper, we focus on the problem of category-level object pose estimation, which is challenging due to the large intra-category shape variation. 3D graph convolution (3D-GC) based methods have been widely used to extract local geometric features, but they have limitations for complex shaped objects and are sensitive to noise. Moreover, the scale and translation invariant properties of 3D-GC restrict the perception of an object's size and translation information. In this paper, we propose a simple network structure, the \emph{HS-layer}, which extends 3D-GC to extract hybrid scope latent features from point cloud data for category-level object pose estimation tasks. The proposed HS-layer: 1) is able to perceive local-global geometric structure and global information, 2) is robust to noise, and 3) can encode size and translation information. Our experiments show that the simple replacement of the 3D-GC layer with the proposed HS-layer on the baseline method (GPV-Pose) achieves a significant improvement, with the performance increased by \textbf{14.5\%} on $5^\circ2\text{cm}$ metric and \textbf{10.3\%} on $\text{IoU}_{75}$. Our method outperforms the state-of-the-art methods by a large margin (\textbf{8.3\%} on $5^\circ2\text{cm}$, \textbf{6.9\%} on $\text{IoU}_{75}$) on REAL275 dataset and runs in real-time (50 FPS)\footnote{Code is available: \href{https://github.com/Lynne-Zheng-Linfang/HS-Pose}{https://github.com/Lynne-Zheng-Linfang/HS-Pose}}.
\vspace{-5mm}

\end{abstract}

\section{Introduction}
\begin{figure}
\centering
\includegraphics[width=1\linewidth, trim = 168 70 175 50, clip,]{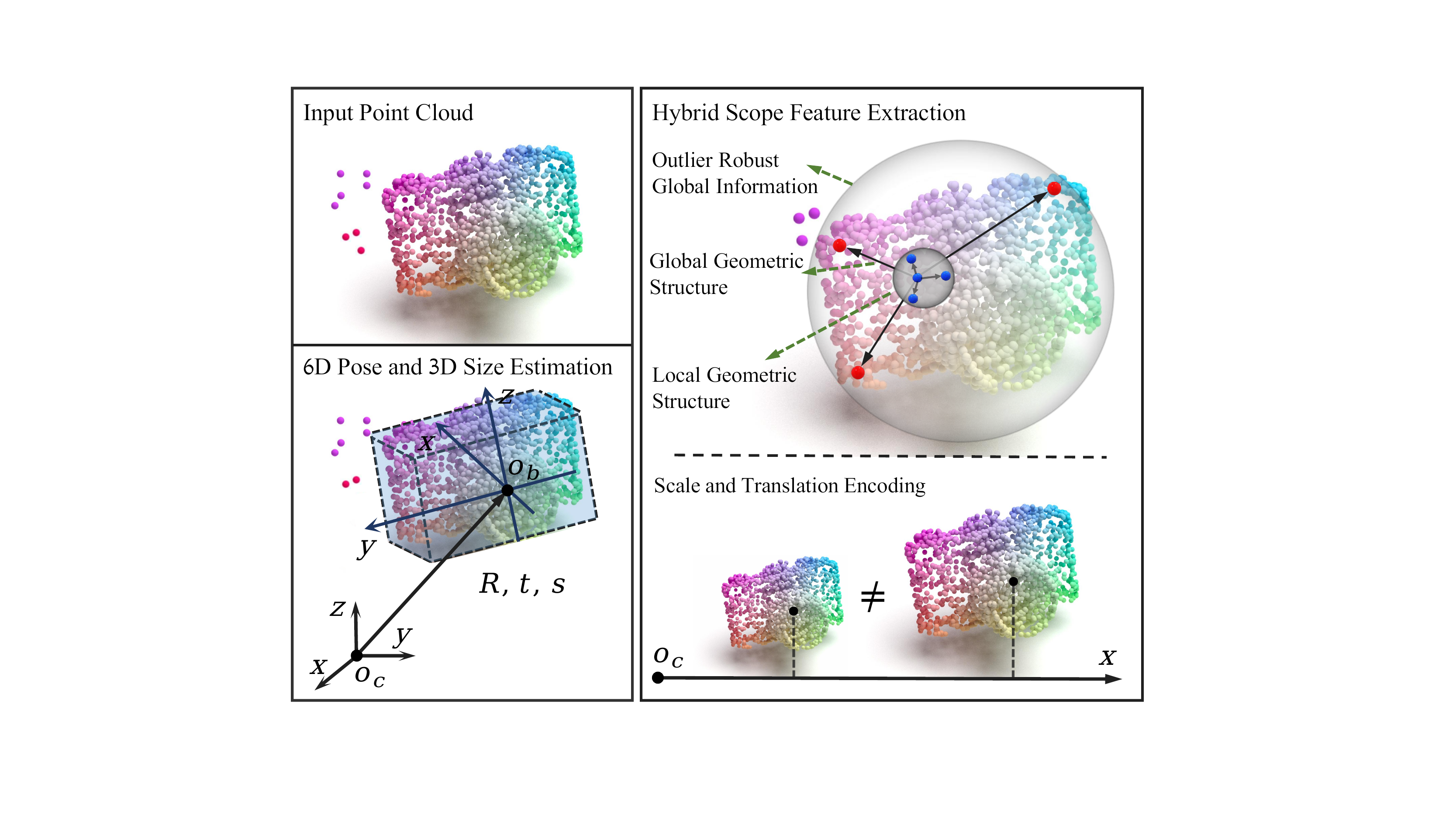}
\vspace{-7mm}
\caption{\footnotesize 
\textbf{Illustration of the hybrid scope feature extraction of the HS-layer}.
As shown in the right figure, the proposed HS-layer possesses various advantages, including the capability of capturing both local and global geometric information, robustness to outliers, and the encoding of scale and translation information. Building upon the GPV-pose, the HS-layer is employed to develop a category-level pose estimation framework, namely \textbf{HS-Pose}. Upon receiving an input point cloud, HS-Pose outputs the estimated 6D pose and 3D size of the object, as shown in the left figure. Given the strengths of the HS-layer, HS-Pose is capable of handling complex object shapes, exhibits robustness to outliers, and achieves better performance compared with existing methods.}

\label{fig:teaser}
\vspace{-3mm}
\vspace{-10px}
\end{figure}             
Accurate and efficient estimation of an object's pose and size is crucial for many real-world applications~\cite{BundleTrack_IROS_2021}, including robotic manipulation~\cite{autonomous_driving_application_2017}, augmented reality~\cite{augmented_reality_application_2021}, and autonomous driving, among others. In these applications, it is essential that pose estimation algorithms can handle the diverse range of objects encountered in daily life.
While many existing works~\cite{Uni6d_2022_CVPR, Epro-PnP_2022_CVPR, ES6D_2022_CVPR, RNNPose_2022_CVPR} have demonstrated impressive performance in estimating an object's pose, they typically focus on only a limited set of objects with known shapes and textures, aided by CAD models. In contrast, category-level object pose estimation algorithms~\cite{category_level_2, CAPTRA_2021, 6D_pack_2020_ICRA, DualPoseNet_2021_ICCV, GPV-Pose_2022_CVPR} address all objects within a given category and enable pose estimation of unseen objects during inference without the target objects' CAD models, which is more suitable for daily-life applications. However, developing such algorithms is more challenging due to the shape and texture diversity within each category.

In recent years, category-level object pose estimation research~\cite{FSNET_2021_CVPR, RBP-Pose_ECCV_2022, SSP-Pose_IROS_22} has advanced rapidly by adopting state-of-the-art deep learning methods. \cite{NOCS_2019_CVPR, CASS_2020_CVPR} gain the ability to generalize by mapping the input shape to normalized or metric-scale canonical spaces and then recovering the objects' poses via correspondence matching. Better handling of intra-category shape variation is also achieved by leveraging shape priors~\cite{Tian_ShapePrior_2020_ECCV, SGPA_2021_ICCV, SSP-Pose_IROS_22}, symmetry priors~\cite{SAR-Net_Lin_2022_CVPR}, or domain adaptation~\cite{UDA-COPE_2022_CVPR, Self-DPDN_ECCV_2022}. Additionally, \cite{FSNET_2021_CVPR} enhances the perceptiveness of local geometry, and~\cite{GPV-Pose_2022_CVPR, RBP-Pose_ECCV_2022} exploit geometric consistency terms to improve the performance further. 

Despite the remarkable progress of existing methods, there is still room for improvement in the performance of the category-level object pose estimation. Reconstruction and matching-based methods~\cite{NOCS_2019_CVPR, Tian_ShapePrior_2020_ECCV, UDA-COPE_2022_CVPR} are usually limited in speed due to the time-consuming correspondence-matching procedure. Recently, various methods~\cite{FSNET_2021_CVPR, SAR-Net_Lin_2022_CVPR, GPV-Pose_2022_CVPR, SSP-Pose_IROS_22, RBP-Pose_ECCV_2022} built on 3D graph convolution (3D-GC)~\cite{3dgcn_lin_CVPR_2020} have achieved impressive performance and run in real-time. They show outstanding local geometric sensitivity and the ability to generalize to unseen objects. However, only looking at small local regions impedes their ability to leverage the global geometric relationships that are essential for handling complex geometric shapes and makes them vulnerable to outliers. In addition, the scale and translation invariant nature of 3D-GC restrict the perception of object size and translation information.

To overcome the limitations of 3D-GC in category-level object pose estimation, we propose the hybrid scope latent feature extraction layer (HS-layer), which can perceive both local and global geometric relationships and has a better awareness of translation and scale. Moreover, the proposed HS-layer is highly robust to outliers. To demonstrate the effectiveness of the HS-layer, we replace the 3D-GC layers in GPV-Pose~\cite{GPV-Pose_2022_CVPR} to construct a new category-level object pose estimation framework, HS-pose. This framework significantly outperforms the state-of-the-art method and runs in real time. Our approach extends the perception of 3D-GC to incorporate other essential information by using two parallel paths for information extraction. The first path encodes size and translation information (STE), which is missing in 3D-GC due to its invariance property. The second path extracts outlier-robust geometric features using the receptive field with the feature distance metric (RF-F) and the outlier-robust feature extraction layer (ORL).

The main contribution of this paper is as follows:
\vspace{-2mm}
\begin{itemize}
    \item We propose a network architecture, the hybrid scope latent feature extraction layer (HS-layer), that can simultaneously perceive local and global geometric structure, encode translation and scale information, and extract outlier-robust feature information. Our proposed HS-layer balances all these critical aspects necessary for category-level pose estimation. 
    \vspace{-2mm} 
    \item We use the HS-layer to develop a category-level pose estimation framework, HS-Pose, based on GPV-Pose. The HS-Pose, when compared to its parent framework, has an advantage in handling complex geometric shapes, capturing object size and translation while being robust to noise.
    \vspace{-2mm}
    \item We conduct extensive experiments and show that the proposed method can handle complex shapes and outperforms the state-of-the-art methods by a large margin while running in real-time (50FPS).                                
\end{itemize}
\vspace{-1mm}

\section{Related Works} \label{sec: related_works}
\textbf{Instance-level object pose estimation}
Instance-level object pose estimation estimates the pose of known objects with the 3D CAD model provided. Existing methods usually achieve the pose using end-to-end regression \cite{SSD_6D_ICCV_2017, CosyPose_ECCV_2020, DeepIM_IJCV_Version_Li_2020}, template matching \cite{Temp6D_CVPR_22, OSOP_2022_CVPR, OVE6D_2022_CVPR}, or 2D-3D correspondence-matching~\cite{Tremblay_2018_Bbox_Corners, OneShot_2d3d_2022_CVPR, Epro-PnP_2022_CVPR, SurfEmb_2022_CVPR, SLAM6D_2022_CVPR, PVN3D_CVPR_2019}. End-to-end regression-based methods estimate object pose directly from the visual observations and have a high inference speed. Template matching methods recover the object pose by comparing the visual observation and usually exhibit robustness to textureless objects. \cite{Hinterstoisser_2016_PPF, Vidal_matching_PPF_2018} use the 3D models as templates, which achieve high accuracy but suffer from low matching speed. In recent years, latent feature-based template matching methods \cite{AAE_perspective, MP-Encoder_CVPR_2020, zheng2022TPAE, PoseRBPF} have achieved real-time performance and have gained popularity. 2D-3D correspondence matching-based methods \cite{ZebraPose_2022_CVPR, DPOD_2019_ICCV} first estimate the 2D-3D correspondences and then retrieve the objects' pose by PnP methods. They show outstanding results for textured objects. The correspondences can be sparse bounding box corners \cite{BB8_Rad_2017_ICCV, SSS_6D_2018_CVPR}, or distinguishable points on the object's surface \cite{Pix2Pose_2019_ICCV, CDPN, PVNet}. While the aforementioned methods have shown impressive capabilities in estimating object pose, their applicability is limited to a few objects and usually needs the corresponding CAD models.

\textbf{Category-level object pose estimation}
Category-level methods estimate the pose of unseen objects within specific categories~\cite{DualPoseNet_2021_ICCV, ShaPO_ECCV_2022, CPSpp_2020, category_level_2}.
NOCS~\cite{NOCS_2019_CVPR} suggests mapping the input shape to a normalized canonical space (NOCS) and retrieving the pose by point matching. \cite{ShaPO_ECCV_2022, CASS_2020_CVPR, UDA-COPE_2022_CVPR} enhance NOCS using a shape prior \cite{Tian_ShapePrior_2020_ECCV}, mapping the shape to a metric scale space \cite{CASS_2020_CVPR}, or domain adaptation \cite{UDA-COPE_2022_CVPR}. \cite{SGPA_2021_ICCV, Self-DPDN_ECCV_2022} leverage structural similarity between the shape prior and the observed object. 
TransNet~\cite{TransNet_ECCV_2022} extends the targets to transparent objects. However, they show limited speed and are unsuitable for real-time applications. CATRE~\cite{liu_2022_catre} explored real-time pose refinement for pose estimation. FS-Net~\cite{FSNET_2021_CVPR} explored local geometric relationships using 3D-GC~\cite{3dgcn_lin_CVPR_2020}, which shows robustness to rotation estimation and runs in real-time. \cite{GPV-Pose_2022_CVPR, SAR-Net_Lin_2022_CVPR, SSP-Pose_IROS_22, RBP-Pose_ECCV_2022} inherit the utilization of 3D-GC and enhance the pose estimation performance in different ways. SAR-Net~\cite{SAR-Net_Lin_2022_CVPR} proposes shape alignment and symmetry-aware shape reconstruction. GPV-Pose~\cite{GPV-Pose_2022_CVPR} presents geometric-pose consistency terms and point-wise bounding box (Bbox) voting. \cite{SSP-Pose_IROS_22, RBP-Pose_ECCV_2022} further enhance \cite{GPV-Pose_2022_CVPR} by shape deformation~\cite{SSP-Pose_IROS_22} and residual Bbox voting~\cite{RBP-Pose_ECCV_2022}. Nonetheless, they only look at local geometric relationships and are limited in handling more complex shapes.

\section{Methodology}
\begin{figure*}[ht]
\centering
\includegraphics[width=1\linewidth]{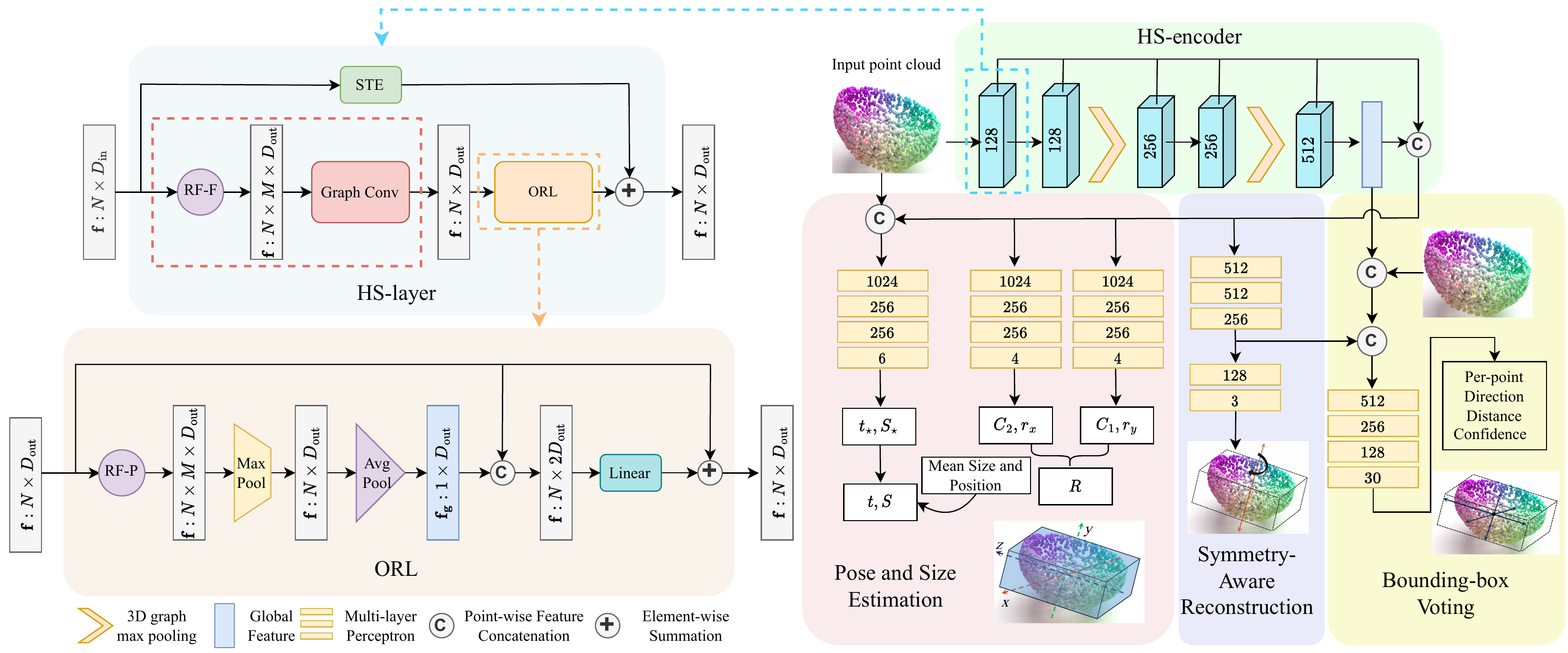}
\vspace{-7mm}
\caption{\footnotesize \textbf{Overview of the proposed HS-Pose.} The core unit of our framework is HS-layer, which extracts the hybrid scope features of the input data in two paths to gain scale and translation encoding and capture outlier robust geometric features. We stack HS-layers and 3D graph max pooling layers to form an HS-encoder, and then connect it to three sub-modules to form HS-Pose. The three sub-modules are used for pose regression, symmetry-aware point cloud reconstruction, and bounding box voting, respectively.} 
\label{fig:framework}
\vspace{-5mm}
\end{figure*} 
This paper considers the category-level pose estimation problem of estimating the 6D pose and 3D size of an arbitrary instance in the same category based on visual observation. In particular, our approach estimates the 3D rotation $\rotation \in SO(3) $, the 3D translation $\translation \in \realR^3$, and the size $\size \in \realR^3$ of object instances based on a depth image, the objects' categories, and segmentation masks. The segmentation mask and category information can be generated by object detectors (\eg MaskRCNN\cite{Maskrcnn_2017}). We use point cloud data $\pointCloud \in \realR^{N\times3}$ as the direct input of our network, which is achieved by back-projecting the segmented depth data and downsampling.

Due to the fact that geometric features are essential for determining an object's pose across different shapes, the 3D graph convolution (3D-GC) \cite{3dgcn_lin_CVPR_2020} is widely adopted in recent category-level object pose estimation methods \cite{FSNET_2021_CVPR, GPV-Pose_2022_CVPR, SSP-Pose_IROS_22, RBP-Pose_ECCV_2022, SAR-Net_Lin_2022_CVPR}. In particular, GPV-Pose\cite{GPV-Pose_2022_CVPR} uses a 3D-GCN encoder, formed by 3D-GC layers, together with geometric consistency terms for category-level object pose estimation and achieves state-of-the-art performance.
However, 3D-GC cannot perceive global geometric features, limiting its capability to handle complex geometric shapes and being sensitive to noise. Also, it is invariant to scale and translation, which contradicts category-level pose estimation tasks (\ie, size and translation estimation).

In this paper, we propose the hybrid scope geometric feature extraction layer (HS-layer) which is based on 3D-GC and keeps its local geometric sensitivity while extending it to have the following characteristics: 1) perception of global geometric structural relationships, 2) robustness to noise, and 3) encoding of size and translation information, particularly for category-level object pose estimation tasks. 

\subsection{Background of 3D-GC}\label{sec:3dgc}
The core unit of 3D-GC is a deformable kernel that generalizes the convolution kernel used in 2D image processing to deal with unstructured point cloud data. In particular, a 3D-GC kernel $K^S$ is defined as:
\begin{equation}
    K^S = \{(\kernelPoint_C, \weight_C), (\kernelPoint_1, \weight_1), \dots, (\kernelPoint_S, \weight_S)\},
\end{equation}
where $S$ is the total number of support vectors, $\kernelPoint_C=[0, 0, 0]^T$ is the central kernel point, $\{\kernelPoint_s\in \realR^3\}_{s=1}^S$ are the support kernel vectors and ${\weight}$ is the weight associated with each kernel vector. The 3D-GC kernel performs a convolution on the receptive field $R^M(\point_i)$, which is the point along with its neighbors and their associated features $\feature$:
\begin{equation}
    \label{eq:receptive_field}
    R^M(\point_i) = \{(\point_i, \feature_i), (\point_m, \feature_m) | \point_m \in \neighbor^M(\point_i)\}.
\end{equation}
Here $\neighbor^M(\point_i)$ is the set of the $M$ nearest neighbor points of $\point_i$. In particular, in \cite{3dgcn_lin_CVPR_2020} the receptive field with point distance metric (RF-P) is used for finding which of the nearest neighbors is within the point distance metric:
\begin{equation}
  \distance_p(\point_i, \point_j) = \norm{\point_i - \point_j}.
\end{equation}

For more details, the readers can refer to the original work \cite{3dgcn_lin_CVPR_2020}. It should be noted that 3D-GC has size and translation invariance by design. Although this invariance may be benefit tasks like segmentation and classification, it harms the pose estimation task as the size and translation are the targets to estimate. 
\subsection{Overall Framework}
The overview of the framework, HS-Pose, is shown in Figure~\ref{fig:framework}. We use the proposed HS-layer to form an encoder (HS-encoder) to extract the hybrid scope latent features from the input point cloud data. 
Then, the extracted latent features are fed into the downstream branches for object pose estimation. 
To demonstrate the effectiveness of the proposed HS-layer, which can be inserted into any category-level object pose estimation method, we construct our hybrid scope pose estimation network (HS-Pose) based on the state-of-the-art 3D-GC based GPV-Pose with minimal modification. Specifically, we only replace the 3D-GC layers of the 3D-GCN encoder of GPV-Pose with the HS-layer and keep all the other settings the same as the original GPV-Pose, which include network layers, network connection structure, and the downstream branches. Therefore, the extracted features from the encoder along with the input point cloud are fed into three modules for object pose regression, symmetric-based point cloud reconstruction, and bounding box voting. During inference, only the encoder and the pose regression module are used.

Inside the HS-layer, we extract the hybrid scope latent features of the input using two parallel paths. The first path performs scale and translation encoding (STE), which provides essential information for size and translation estimation. The second path extracts outlier-robust geometric features by leveraging local and global geometric relationships, as well as global information in two phases. In the first phase, we form the receptive fields of points based on their feature distances (RF-F), then feed them to a graph convolution (GC) layer to extract high-level geometric features. The output of the GC layer is taken as the second phase's input and passes through an outlier-robust feature extraction layer (ORL), where each point feature is adjusted by an outlier robust global information. 
The final output of the HS-layer is the element-wise summation of the features of both paths.

\subsection{Scale and Translation Encoding (STE)}
As mentioned earlier, even though 3D-GC provides geometric features crucial in rotation estimation, it loses the essential translation and scale information necessary for pose estimation. To address this problem, existing 3D-GC-based methods try to use another network for translation and size estimation \cite{FSNET_2021_CVPR} or concatenate the point cloud data with the extracted features for downstream estimation tasks with the assistance of other modules (\ie, bounding box voting)\cite{GPV-Pose_2022_CVPR, SSP-Pose_IROS_22, RBP-Pose_ECCV_2022}. While these methods are effective and all achieve improvements from the baseline, we emphasize the scale and translation information is beneficial during the latent feature extraction phase. 

As shown in Figure~\ref{fig:framework}, our suggestion is to connect in parallel a linear layer (see STE in HS-layer in the figure) to the geometric extraction path and then perform element-wise summation for their output features:
\begin{equation}
    \outputFeature_n = \geometricLayer(\feature_n) + \linearLayer(\feature_n),
\end{equation}
where $\linearLayer$ and $\geometricLayer$ apply linear transformation and geometric feature extraction on the features of the points, respectively, and $\feature_n$ is the $n$-th point's feature. In particular, we use the points' positions for size and translation encoding in the first layer since there are no features in the original point cloud. Our ablation study in Table~\ref{tbl:ablation_full} shows that this design choice keeps the advantage of geometric feature extraction, and boosts the performance of translation and scale estimation.


\subsection{Receptive field with feature distance (RF-F)}
\begin{figure}
\centering
\includegraphics[width=0.97\linewidth, trim = 160 66 120 230, clip,]{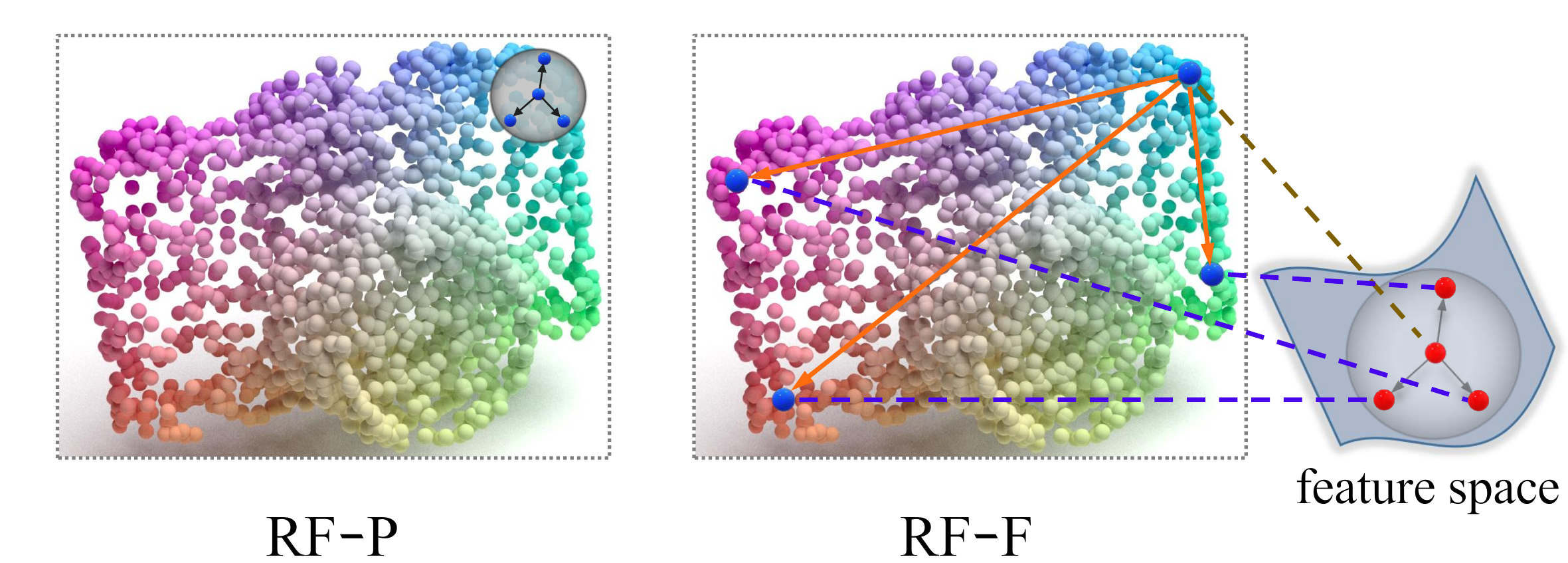}
\vspace{-2mm}
\caption{\footnotesize \textbf{The illustration and comparison of the receptive field between RF-P and RF-F.} RF-P could only capture geometric structures in a small local region, while RF-F could capture more complex global geometric relationships among the latent features for each point in a high-dimensional hyperspace. }
\label{fig:reception_field}
\vspace{-15px}
\end{figure}    
As introduced in Sec.~\ref{sec:3dgc}, 3D-GC learns awareness of local geometric features by forming receptive fields with point Euclidean distance metric (RF-P) and then using the deformable kernel-based graph convolution to extract geometric features for the receptive fields. However, RF-P restricts the perception to small local regions. Even though the perceived regions can be enlarged when cooperating with 3D graph pooling, it can not perceive the global geometric relationships essential for complex geometric structures. This limitation is also exhibited in the performance of category-level object pose estimation tasks \cite{GPV-Pose_2022_CVPR}, where the methods show impressive capability in handling simple geometric shapes (\eg bowl) while encountering difficulty with more complex shapes (\eg mug and camera). However, this limitation has not been well addressed. To this end, we extend the 3D-GC and propose a simple manner to leverage global geometric structural relationships.

We suggest forming the receptive field with the feature distance metric (RF-F). Specifically, we find $\point_i$'s neighbors using the feature distance metric:
\begin{equation}
\distance_f(\point_i, \point_m) = \norm{\feature_i -  \feature_m}.  
\end{equation} In other words, with the feature distance metric, the distance between two points is the Euclidean distance between their associated features. We denote the corresponding receptive fields as $R_{f}^M(\point_i)$. 
This receptive field has the advantage that it is not restricted to local regions; distant points with similar features can also be included.

Figure~\ref{fig:reception_field} shows the difference between RF-P and RF-F. RF-F can capture a larger receptive field and, therefore, can capture geometric relationships in a larger area, while the RF-P always formed with local regions. 
For initialization, in the first layer, we use RF-P and set all the features $\feature$ to 1. The RF-F is used in the following layers for extracting higher-level geometric relationships. 


\subsection{Outlier robust feature extraction layer (ORL) }
\begin{figure}
\centering
\includegraphics[width=0.8\linewidth, trim = 180 130 150 145, clip,]{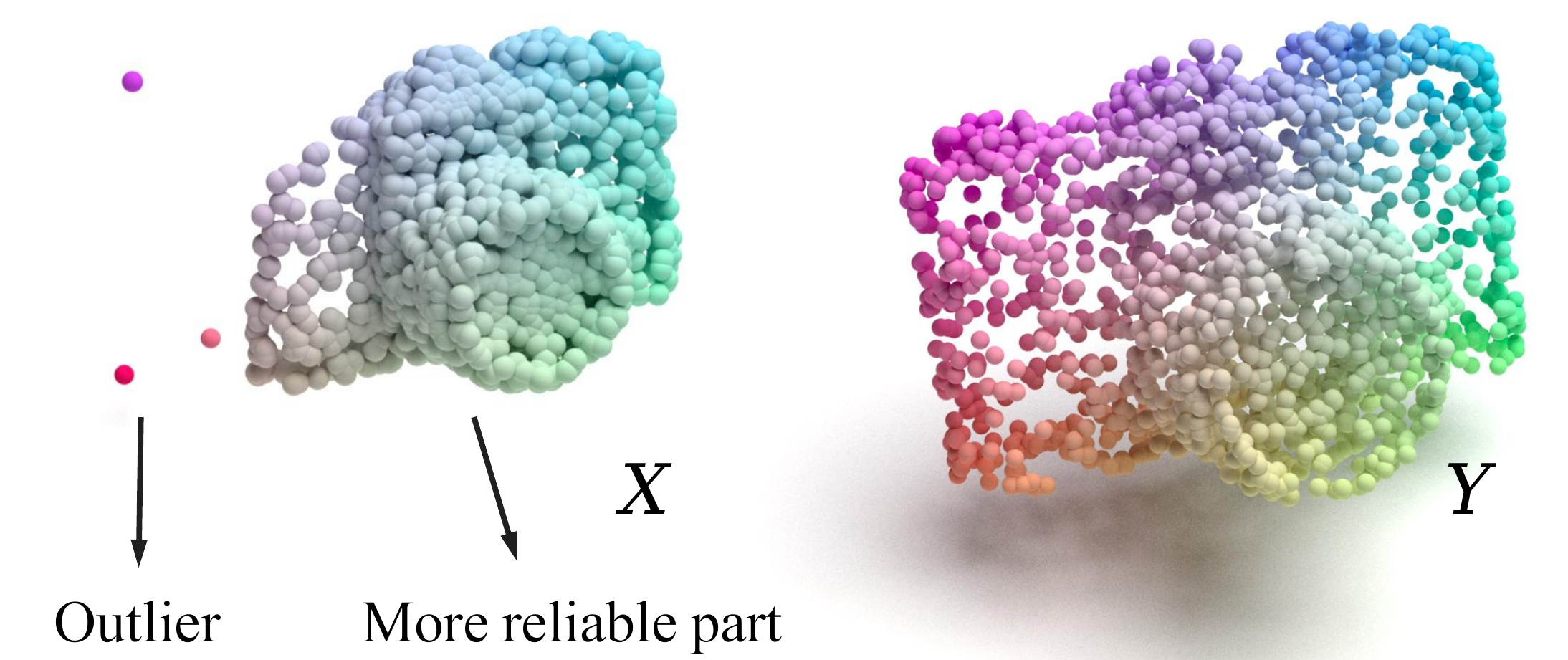}
\vspace{-3mm}
\caption{\footnotesize \textbf{The design intuition of the outlier robust feature extraction layer (ORL).} $X$ is the input point cloud of a camera with outliers, and $Y$ is the complete shape. Having a perception of global information, especially the more reliable part, helps the network gain resistance to noise.}
\label{fig:noise_illus}
\vspace{-15px}
\end{figure}              
3D-GC's sensitivity to noise influences the category-level methods \cite{FSNET_2021_CVPR, GPV-Pose_2022_CVPR, RBP-Pose_ECCV_2022, SSP-Pose_IROS_22} that are based on it. 
To address this problem, we introduce an outlier robust feature extraction layer (ORL) on top of the 3D-GC layer, which enhances the method's robustness to noise.
The ORL is constructed as follows. Denote the input to this layer as $\{(\point_1, \feature_1), \dots, (\point_N, \feature_N)\}$, where $\feature_n \in \realR^D$ is the feature of point $\point_n$. 
As illustrated in Figure~\ref{fig:noise_illus}, outliers are distractive, and their features $\feature$ should not be trusted. To focus on the global information of the more reliable part, we need a mechanism to alleviate the deviation caused by the outliers. Using the global average or maximum pooling directly is limited in addressing this, as all points are taken equally in the pooling procedure.

To lower outliers' influence, we propose using the local region as a guide to extract the global feature.
As shown in Figure~\ref{fig:framework} (see ORL), we first use RF-P to find the $M$ nearest neighbors of each point $\neighbor_p^M(\point_n)$. Then, we extract the channel-wise max features of $\neighbor_p^M(\point_n)$ using a maximum pooling layer. It should be noted that the points in the reliable parts are more likely to be presented in other points' receptive fields and thus contribute more to the results of the max pooling. The output of the max pooling layer is then passed to a global average pooling layer to get the global feature $\globalFeature$. We then generate an adjusting feature using the $\globalFeature$ and the original input per-point feature $\feature_n$ by first concatenating them and then feeding them to a linear layer. The final output of ORL is the result of the summation of the adjusting feature and the input features $\feature_n$ of this layer. 




\section{Experiments}
\begin{table*}
\begin{center}

\caption{\textbf{Ablation studies on REAL275.} }
\vspace{-4mm}
\caption*{\footnotesize Higher score means better performance. Overall best results are in bold, and the second-best results are underlined.}
\label{tbl:ablation_full}
\vspace{-3mm}
\resizebox{1\linewidth}{!}
{\footnotesize

\begin{tabular}{@{}c|l|ccc|cccc|cc|c@{}}
\toprule

Row & Method &$\text{IoU}_{25}$   &$\text{IoU}_{50}$    &$\text{IoU}_{75}$     &$5^\circ 2$cm &$5^\circ 5$cm &$10^\circ 2$cm &$10^\circ 5$cm &$2$cm& $5^\circ$ &Speed(FPS)  \\
\midrule            
\midrule            

A0 & GPV-Pose\cite{GPV-Pose_2022_CVPR} (baseline)  &{84.2} &\textbf{83.0} &64.4 &32.0 &42.9 &55.0 &73.3 &69.7 &44.7 &\textbf{69}\\

\midrule
B0 & A0 + STE   &{84.2} &{82.2} &{73.1} & {36.4} &{45.1} &{62.2} &{76.7} &75.6 &47.4 &\underline{66}\\ 

B1 & A0 + RF-F   &{84.2} &\underline{82.8} &{67.7} & {38.9} &{52.3} &{62.1} &{81.8} &71.7 &{56.1} &65\\

B2 & A0 + STE + RF-F   &{84.1} &{82.0} &{72.0} & {42.7} &{53.7} &{63.4} &{79.2} &75.7 &{57.0} &64\\ 
\midrule

C0 & A0 + STE + RF-F + Average Pool   &{84.1} &{81.7} &{73.4} & {43.7} &{54.8} &{65.7} &{81.6} &75.7 &\underline{58.5} &62\\ 

C1 & A0 + STE + RF-F + Max Pool   & {84.2} &{81.7} &\underline{74.8} & {44.3} &{54.5} &{66.9} &{81.8} &{77.3} &{58.1} &{62}\\ 

\midrule
\textbf{D0} & A0 + STL + RF-F + ORL (\textbf{Full})  &{84.2} &{82.1} &{74.7} &\textbf{46.5} &\underline{55.2} &\underline{68.6} &\underline{82.7} &\textbf{78.2} &{58.2} &50\\ 

\midrule
E0 & D0: Neighbor number: 10 $\rightarrow$ 20  &\textbf{84.3} &\underline{82.8} &\textbf{75.3} &\underline{46.2} &\textbf{56.1} &\textbf{68.9} &\textbf{84.1} &\underline{77.8} &\textbf{59.1} &38\\ 
\bottomrule
\end{tabular}
}
\vspace{-7mm}
\end{center}
\end{table*}

\textbf{Implementation details:}
To rigorously verify the effectiveness of the proposed HS-layer and ensure a fair comparison with the baseline GPV-Pose, we construct the HS-Pose by replacing GPV-Pose's 3D-GC layer with the HS-layer while keeping the overall network structure and network parameters identical to the GPV-Pose, as shown in Figure~\ref{fig:framework}. For a fair comparison, we choose 10 neighbors for the RF-F, consistent with the RF-P in GPV-Pose. The neighbor number of ORL is the same as the RF-F. No other parameters need to be set for the HS-layer as they only depend on the input and output. We also keep the settings, data augmentation strategy, loss terms, and their parameters, the same as those in GPV-Pose's official code\footnote{\label{foot:gpv_cite}\href{https://github.com/lolrudy/GPV_Pose}{https://github.com/lolrudy/GPV\_Pose}}. Following GPV-Pose, the off-the-shelf object detector MaskRCNN~\cite{Maskrcnn_2017} is employed to generate instance segmentation masks, and 1028 points are randomly sampled as the input to the network. The code is developed using \texttt{PyTorch}. We run all experiments on a computer equipped with an Intel(R) Core(TM) i9-10900K CPU, 32 GB RAM, and an NVIDIA GeForce RTX 3090 GPU. All categories are trained together with a batch size of 32, and the training epochs are set to 150 and 300 for REAL275 and CAMERA25 datasets, respectively. The Ranger optimizer\cite{ranger_optimizer_1_ICLA_2019, Ranger_optimizer_2_ECCV_2020, Ranger_optimizer_3_2019_NIPS} is used with the learning rate starting at $1e^{-4}$ and then decreasing based on a cosine schedule for the last $28\%$ training phase.

\textbf{Baseline methods: }
We use GPV-Pose~\cite{GPV-Pose_2022_CVPR} as the baseline for the ablation study. Since GPV-Pose did not provide the performance of $10^\circ2\text{cm}$, $2\text{cm}$, and $5^\circ$, we generate them using their official code\footnoteref{foot:gpv_cite}. To ensure a fair comparison of their relative speeds, we report GPV-Pose's speed on our machine using the same evaluation code as ours. The results of the other methods are taken directly from the corresponding papers. 

\textbf{Datasets: }
We evaluate our method on REAL275~\cite{NOCS_2019_CVPR} and CAMERA25~\cite{NOCS_2019_CVPR}, the two most popular benchmark datasets for category-level object pose estimation. REAL275 is a real-world dataset that provides 7k RGB-D images in 13 scenes. It contains 6 categories of objects (can, laptop, mug, bowl, camera, and bottle), and every category contains 6 instances. The training data comprises 4.3k images from 7 scenes, with 3 objects from each category shown in different scenes. The testing data includes 2.7k images from 6 scenes and 3 objects from each category. CAMERA25 is a synthetic RGB-D dataset that contains the same categories as REAL275. It provides 1085 objects for training and 184 for testing. The training set contains 275K images, and the testing set contains 25K.

\textbf{Evaluation metrics:} 
Following~\cite{RBP-Pose_ECCV_2022, GPV-Pose_2022_CVPR}, we use the mean average precision (mAP) of the \textit{3D Intersection over Union (IoU)} with thresholds of $25\%$, $50\%$, and $75\%$ to evaluate the object's size and pose together. We evaluate the rotation and translation estimation performance using the metrics of $5^\circ$, $10^\circ$, $2\text{cm}$ and $5\text{cm}$, which means an estimation is considered correct if its corresponding error is lower than the threshold. The pose estimation performance is also evaluated using the combination of rotation and translation thresholds: $5^\circ2\text{cm}$, $5^\circ5\text{cm}$, $10^\circ2\text{cm}$, and $10^\circ5\text{cm}$.

\subsection{Ablation Study}
To validate the proposed architecture, we conduct intensive ablation studies using the REAL275~\cite{NOCS_2019_CVPR} dataset. We incrementally add the proposed strategies (STE, RF-F, and ORL) on the baseline (GPV-Pose) to study their influences. The full ablation study results are shown in Table~\ref{tbl:ablation_full}.

\textbf{[AS-1] Scale and translation encoding (STE).}
To demonstrate the effectiveness of STE and highlight the significance of scale and translation awareness when extracting latent features, we parallelly connected a single linear layer to each 3D-GC layer in the encoder of the GPV-Pose. The results in Table~\ref{tbl:ablation_full}, specifically the [B0] row, indicate that the inclusion of STE has a significant positive impact on scale and translation estimation ($\textbf{8.7\%}$ improvement on $\text{IoU}_{75}$ and $\textbf{5.9\%}$ improvement on $2\text{cm}$) while also slightly improving rotation estimation ($2.7\%$ improvement on $5^\circ$).
As shown in Table~\ref{tbl:cat_perform}, such a simple addition even outperforms the SSP-Pose in several strict metrics ($\IoUSevenFive$, $\fiveDtwoC$, and $\fiveDfiveC$) and shows a notable improvement of $6.8\%$ on the $\IoUSevenFive$ metric, despite that the SSP-Pose extends the GPV-Pose using a much more complex shape deformation module. The experiment results demonstrate the effectiveness of STE.

\textbf{[AS-2] Receptive field with feature distance (RF-F).}
To show the usefulness of the proposed RF-F strategy and to demonstrate the importance of the global geometric relationships, we apply RF-F on GPV-Pose. From the results in Table~\ref{tbl:ablation_full} ([B1]), we see that RF-F has a substantial impact on rotation estimation and brings a performance leap by $\textbf{11.4\%}$ on $5^\circ$ metric. In addition, it improves the performance on $\text{IoU}_{75}$ and $2\text{cm}$ by $3.3\%$ and $2.0\%$, respectively, thanks to the fact that having a sense of the global geometric relationships is helpful in finding the object's center and shape boundary. When comparing the experimental results with the state-of-the-art methods in Table~\ref{tbl:cat_perform}, our simple RF-F strategy achieves 
comparable performance with the state-of-the-art methods and outperforms them on the stricter metrics (\eg $5^\circ2\text{cm}$ and $5^\circ5\text{cm}$).

\textbf{[AS-3] The combination of RF-F and STE.}
To exhibit the benefit of leveraging global geometric relationships and size-translation awareness, we conduct an experiment that combines RF-F and STE. As shown in [B2], the cooperation of RF-F and STE enhances each other and contributes to a better performance than their individual results. When compared with the baseline method, GPV-Pose, the combination of RF-F and STE improves $5^\circ5\text{cm}$ by $\textbf{10.8\%}$, $5^\circ$ by $\textbf{12.3\%}$ and $\text{IoU}_{75}$ by $\textbf{7.6\%}$.

\textbf{[AS-4] Outlier robust feature extraction layer (ORL).}
To demonstrate the effectiveness of the ORL, we add the ORL on top of [AS-3]. The results shown in the [D0] row of Table~\ref{tbl:ablation_full} demonstrate that using global features to adjust per-point feature extraction is helpful for both pose and size estimation with an improvement of ${5.2\%}$ ($10^\circ2\text{cm}$) and ${2.7\%}$ ($\text{IoU}_{75}$), respectively. To check the effectiveness of the outlier robust global feature, we further conduct two experiments by replacing the outlier robust global feature with two popular global pooling methods: average pooling [C0] and max pooling [C1]. The results of [D0], [C0], and [C1] all show the contribution of global information to pose estimation. The comparison between [D0] and [C0, C1] shows that the outlier robust global feature plays a positive role and enhances the overall performance.

\begin{figure}[!t]
\centering
\vspace{1mm}
\includegraphics[width=1\linewidth, trim = 0 0 0 50 , clip]{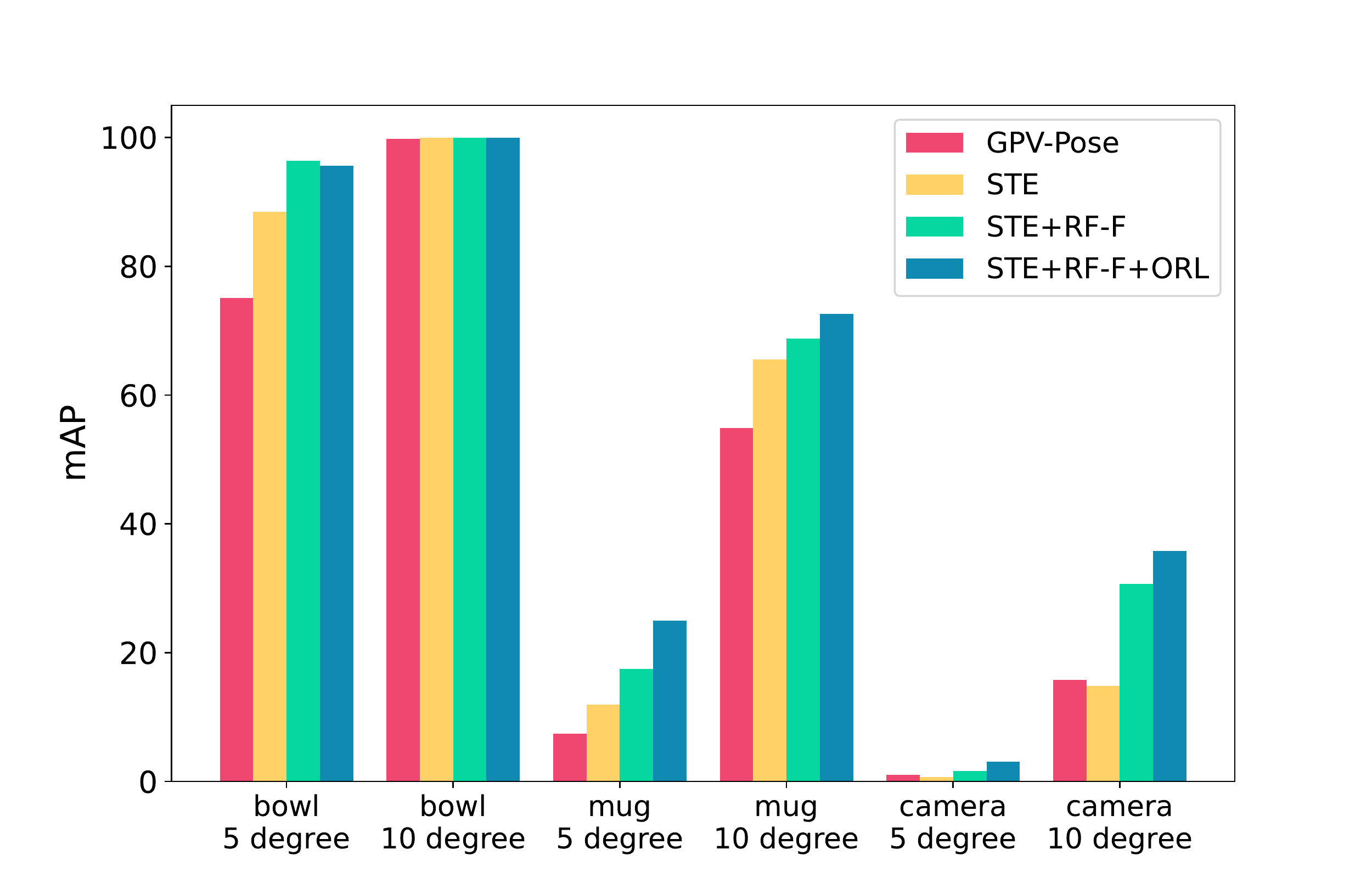}
\vspace{-8mm}
\caption{\footnotesize \textbf{The rotation estimation of the proposed three strategies and GPV-Pose on categories with different geometric complexity.} The figure shows the rotation estimation mAP ($5^\circ$ and $10^\circ$) on objects with different geometric complexities (\ie bottle is the simplest and the camera is the most complex one). Our method boosted the rotation estimation of the simple shape (bowl) to almost $100\%$ and increased the rotation mAP on more complex objects (mug and camera) by a large margin. 
} 
\vspace{-13px}
\label{fig:bar}
\end{figure} 
\textbf{[AS-5] Capability of handling complex shapes.} \label{ab:complex_shape}
To exhibit the proposed method's capability in handling complex geometric shapes, we compare the rotation estimation results of the three proposed strategies (STE, RF-F, and ORL) and GPV-Pose on categories with different shape complexity in Figure~\ref{fig:bar}. As shown in the figure, the proposed method increases the mAP of categories with complex shapes (\ie mug and camera) and handles simple shapes (\ie bowl) with ease. The figure also demonstrates the effectiveness of leveraging global geometric relationships (STE+RF-F \vs STE) and shows the usefulness of outlier robust global information guided feature extraction in ORL (STE+RF-F+ORL \vs STE+RF-F).

\begin{figure}[t]
\centering
\includegraphics[width=1\linewidth, trim = 10 0 675 5, clip]{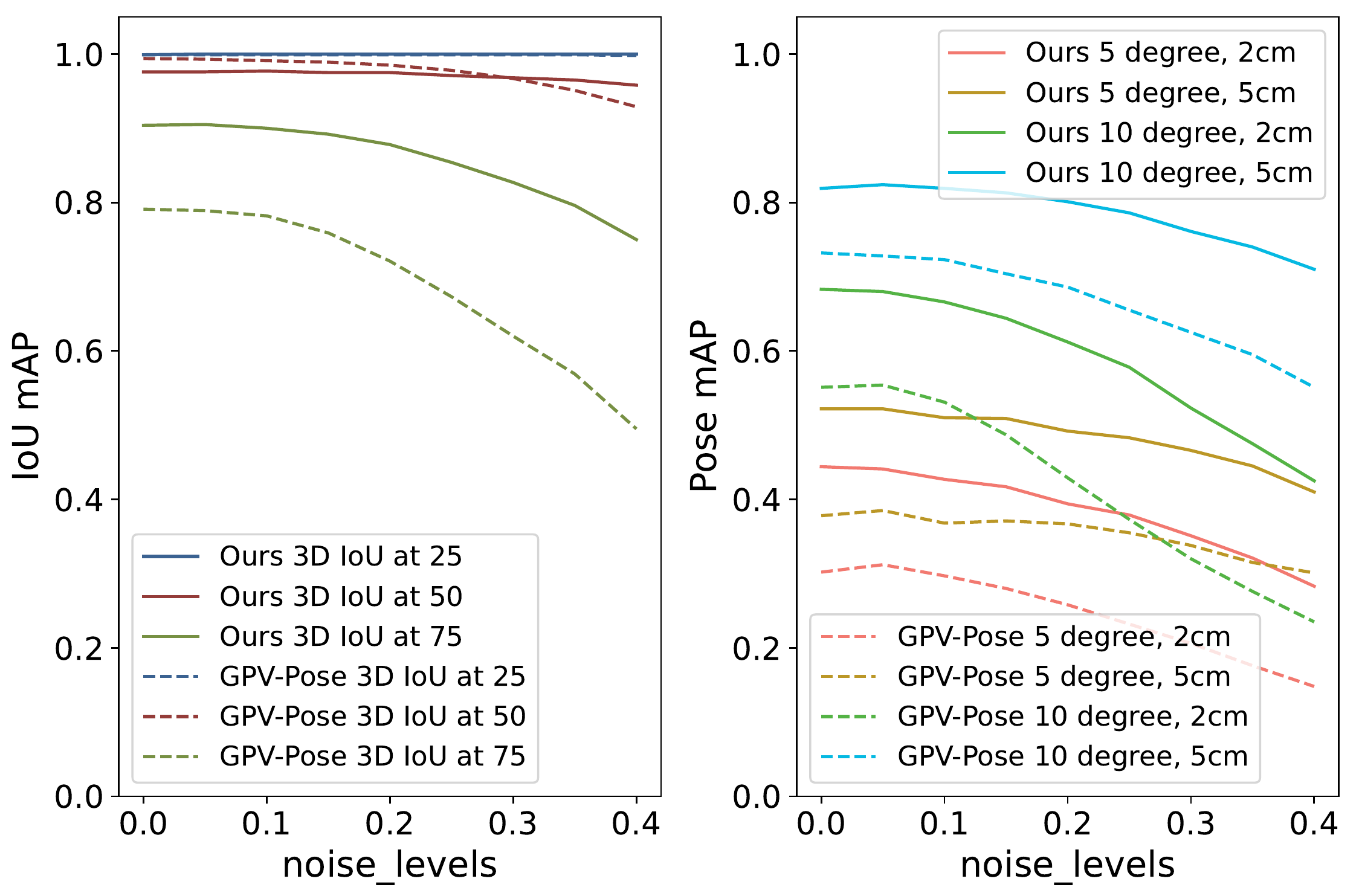}
\vspace{-7mm}
\caption{\footnotesize \textbf{The comparison of noise resistance between GPV-Pose and the proposed HS-Pose under different outlier ratios (from 0.0\% to 40.0\%).} Our method outperforms GPV-Pose by a large margin across all outlier ratio levels and is steadier when the outlier ratio increases.
} 
\vspace{-13px}
\label{fig:noise_resistance}
\end{figure}

\textbf{[AS-6] Noise resistance.}
To demonstrate the outlier robustness of the proposed method, we tested GPV-Pose and our method under different outlier ratios. As shown in Figure~\ref{fig:noise_resistance}, our method outperforms GPV-Pose by a large margin across a range of outlier ratios and is steadier when the outlier ratio increases. More details are in the Supplementary.

\textbf{[AS-7] Neighbor numbers.}
We investigate the influence of neighbor numbers used in ORL and RF-F on the performance. The details are presented in the supplementary. The results show that the performance is best when the neighbor numbers are in a certain range. We also observed that using the same neighbor numbers in ORL and RF-F enhances the performance: the precision results are best when the neighbor numbers for both ORL and RF-F are 20 or 30. The results for 20 neighbor numbers are shown in row [E0] of Table~\ref{tbl:ablation_full}, which outperforms the results with 10 neighbors. It should be noted that, for a fair comparison with GPV-Pose and focusing on the HS-layer's structural design, we use the results with 10 neighbors (as GPV-Pose) in all tables and figures if not specified.

\begin{table*}[!t]
\begin{center}

\caption{\centering\textbf{Comparison with the state-of-the-art methods (depth only) on REAL275 dataset.} }
\vspace{-4mm}
\caption*{\footnotesize Higher score means better performance. Overall best results are in bold, and the second-best results are underlined.}
\label{tbl:cat_perform}
\vspace{-3mm}
\resizebox{0.8\linewidth}{!}
{%
\footnotesize

\begin{tabular}{@{}r|ccc|ccccc|c@{}}
\toprule
Method & $\text{IoU}_{25}$   &$\text{IoU}_{50}$    &$\text{IoU}_{75}$     &$5^\circ 2$cm &$5^\circ 5$cm &$10^\circ 2$cm &$10^\circ 5$cm &$10^\circ 10$cm&Speed(FPS)  \\
\midrule            
\midrule            
SAR-Net \cite{SAR-Net_Lin_2022_CVPR} &- &79.3 &62.4 &31.6 &42.3 &50.3 &68.3 &- &10\\

FS-Net\tablefootnote{We use the result provided by GPV-Net, which is higher than the reported result in the FS-Net paper.} \cite{FSNET_2021_CVPR} &{84.0} &81.1 &63.5 &19.9 &33.9 &- &69.1 &71.0 &20\\

UDA-COPE~\cite{UDA-COPE_2022_CVPR}  &- &79.6 &57.8 &21.2 &29.1 &48.7 &65.9 &- &- \\

SSP-Pose \cite{SSP-Pose_IROS_22}  &{84.0} &\underline{82.3} &66.3 &34.7 &44.6 & - &{77.8} &\underline{79.7} & 25\\ 

RBP-Pose\cite{RBP-Pose_ECCV_2022}  & - & - &\underline{67.8} &\underline{38.2} &\underline{48.1} &\underline{63.1} &\underline{79.2} &- & 25 \\

GPV-Pose \cite{GPV-Pose_2022_CVPR}   &{84.1} &\textbf{83.0}&{64.4}&32.0 &{42.9} &55.0 &{73.3}  &{74.6}&\textbf{69}\\

\midrule
\textbf{Ours} &\textbf{84.2} &{82.1} &\textbf{74.7} &\textbf{46.5} &\textbf{55.2} &\textbf{68.6} &\textbf{82.7} &\textbf{83.7} & \underline{50}\\ 

\bottomrule

\end{tabular}
}
\vspace{-8px}
\end{center}
\end{table*}
\begin{figure*}[htbp]
	\centering
	\begin{minipage}{0.05\linewidth}
	    GPV-Pose:
	\end{minipage}
	\begin{minipage}{0.23\linewidth}
		\centering
		\includegraphics[width=\linewidth]{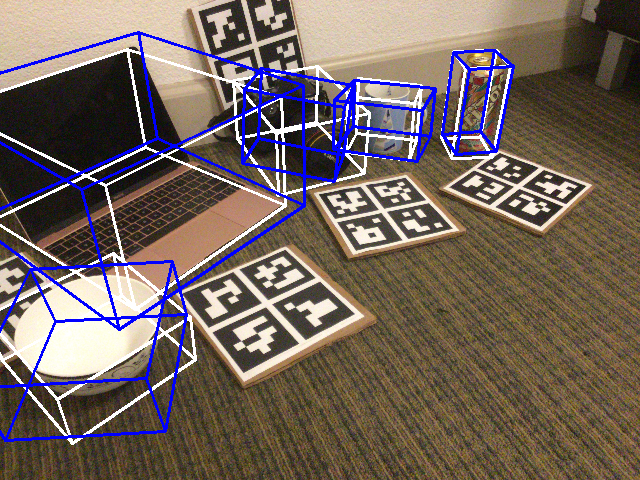}
	\end{minipage}
	\begin{minipage}{0.23\linewidth}
		\centering
		\includegraphics[width=\linewidth]{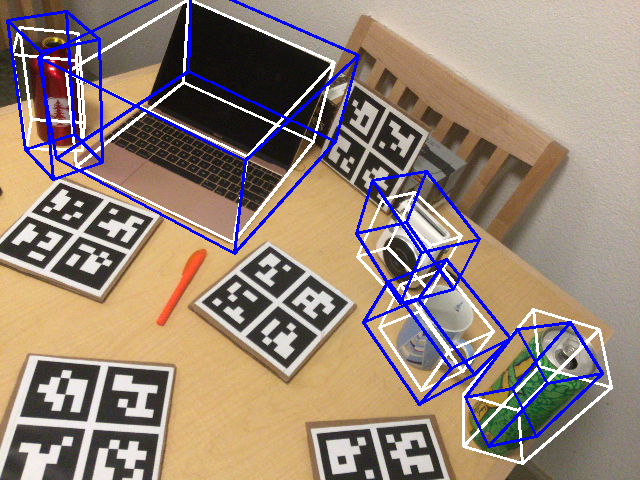}
	\end{minipage}
	\begin{minipage}{0.23\linewidth}
		\centering
		\includegraphics[width=\linewidth]{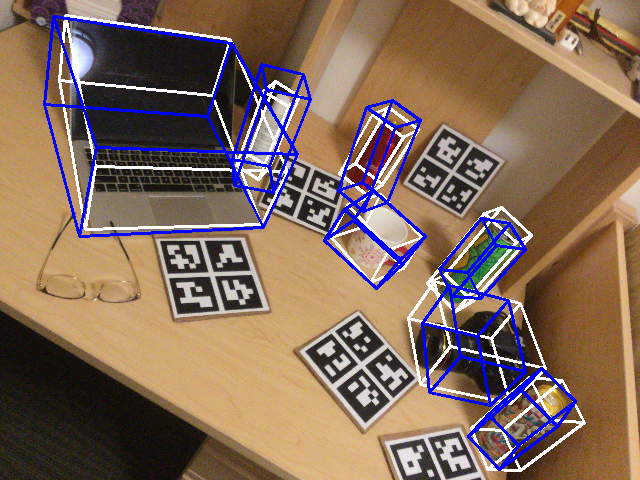}
	\end{minipage}
	\begin{minipage}{0.23\linewidth}
		\centering
		\includegraphics[width=\linewidth]{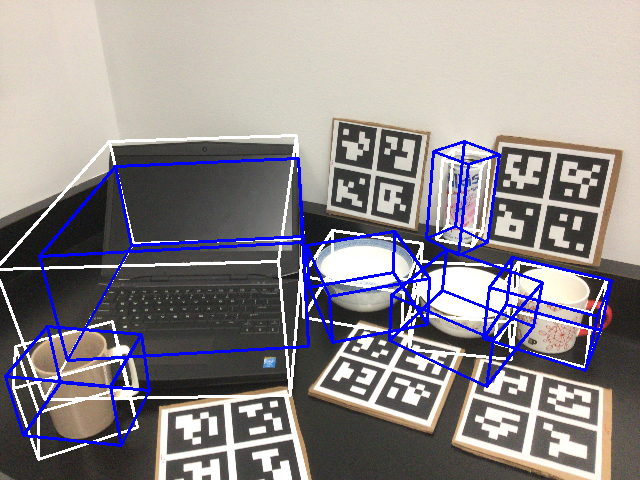}
	\end{minipage}
	
	\begin{minipage}{0.05\linewidth}
	    Ours:
	\end{minipage}
	\begin{minipage}{0.23\linewidth}
		\centering
		\includegraphics[width=\linewidth]{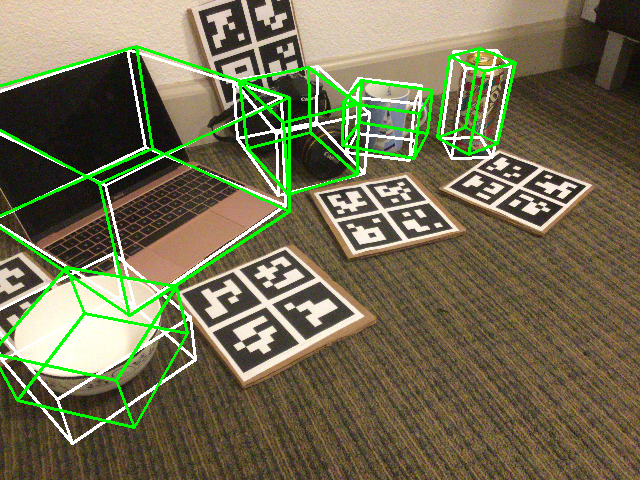}
	\end{minipage}
	\begin{minipage}{0.23\linewidth}
		\centering
		\includegraphics[width=\linewidth]{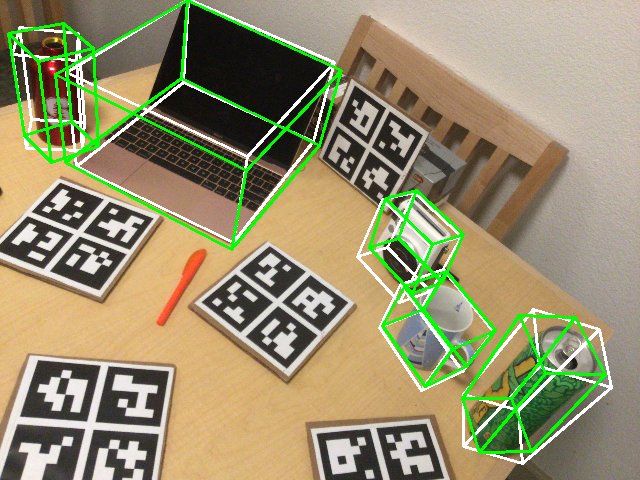}
	\end{minipage}
	\begin{minipage}{0.23\linewidth}
		\centering
		\includegraphics[width=\linewidth]{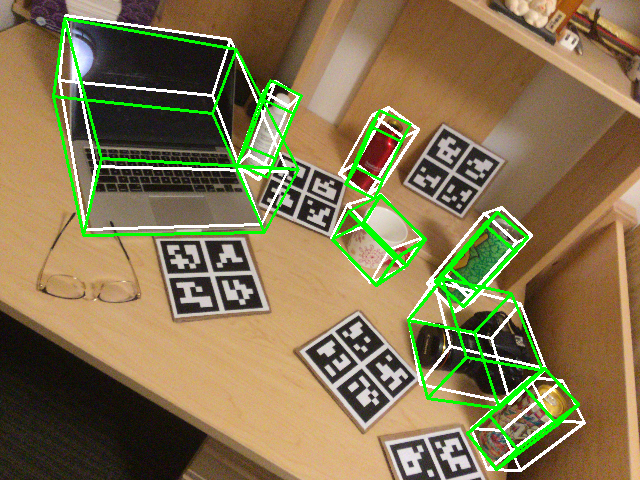}
	\end{minipage}
	\begin{minipage}{0.23\linewidth}
		\centering
		\includegraphics[width=\linewidth]{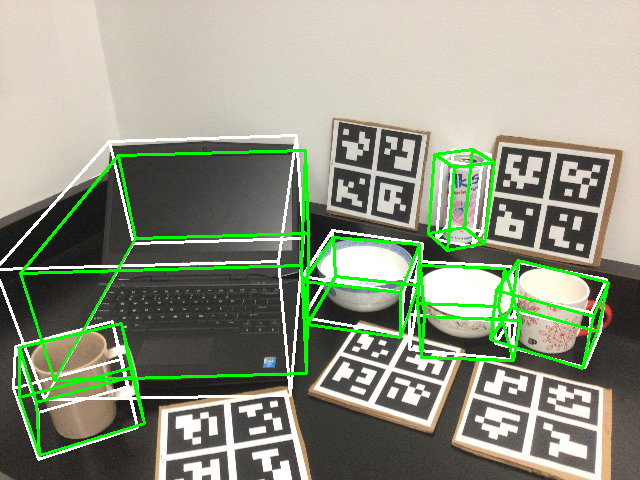}
	\end{minipage}

        \vspace{-2mm}
	\caption{\footnotesize \textbf{Qualitative results of our method (green line) and the GPV-Pose (blue line)}. The ground truth results are shown with white lines. The estimated rotations of symmetric objects (\eg bowl, bottle, and can) are considered correct if the symmetry axis is aligned.}
\label{fig:qualitative}
\vspace{-4mm}
\end{figure*}

\subsection{Comparison With State-of-the-Art Methods}
\textbf{Results on REAL275 dataset: }
We compare the performance of the proposed HS-Pose with the state-of-the-art methods in Table~\ref{tbl:cat_perform}, which shows the mAP scores in different metrics. We choose methods that use depth only for pose estimation for a fair comparison. As shown in the table, our method outperforms the state-of-the-art methods in all metrics except the $\text{IoU}_{50}$ in which our method also have comparable performance. Besides, our method can run in real-time. It is worth noting that our method outperforms the second rank on strict metrics by a large margin, with $\textbf{8.3\%}$ improvement on $5^\circ2\text{cm}$, and $\textbf{7.1\%}$ on $5^\circ5\text{cm}$, and $\textbf{6.9\%}$ on $\text{IoU}_{75}$. We also provide the comparison with methods \cite{Self-DPDN_ECCV_2022, SGPA_2021_ICCV, DualPoseNet_2021_ICCV, Tian_ShapePrior_2020_ECCV, NOCS_2019_CVPR, CASS_2020_CVPR, CR-Net_2021_IROS} and that by using other data modalities (\eg RGB and RGB-D) in the supplementary, we outperform the state-of-the-art on 5 metrics out of 9 and achieved the second rank on 3 metrics. Notably, most of them leverage synthetic data, whose datasets contain many more images and objects for training purposes, and also exhibit a limited inference speed. Our method is trained using REAL275 with only 1.6k images and 18 objects while achieving real-time performance. A qualitative comparison between GPV-Pose and our method is shown in Figure~\ref{fig:qualitative}. Our method achieves a better size and pose estimation (\eg the first three columns), shows robustness to occlusion (\eg the laptop in the last column), and handles complex shapes better (\eg the cameras and mugs in each column). 
\begin{table}[!t]
\begin{center}

\caption{\footnotesize \textbf{Comparison with state-of-the-art methods (depth-only) on CAMERA25 dataset.} Overall best results are in bold, and the second-best results are underlined. \emph{Prior} denotes whether the method uses shape priors.}
\label{tbl:cat_perform_camera}
\vspace{-3mm}
\resizebox{\linewidth}{!}
{%

\begin{tabular}{@{}r|c|cc|cccc@{}}
\toprule

Method  & Prior & $\text{IoU}_{50}$ & $\text{IoU}_{75}$ & $5^\circ 2\text{cm}$ & $5^\circ 5\text{cm}$ & $10^\circ 2\text{cm}$ & $10^\circ 5\text{cm}$    \\ 
\midrule
\midrule            
SAR-Net\cite{SAR-Net_Lin_2022_CVPR}   &\checkmark  & 86.8  & 79.0  & 66.7 & 70.9 & 75.3  & 80.3  \\
SSP-Pose\cite{SSP-Pose_IROS_22}   & \checkmark  & -  & 86.8  & 64.7 & 75.5 & -  & 87.4  \\
RBP-Pose \cite{RBP-Pose_ECCV_2022}    & \checkmark   & 93.1  & \underline{89.0}  & \textbf{73.5} & \underline{79.6} & \textbf{82.1}  & \textbf{89.5}  \\ 
GPV-Pose \cite{GPV-Pose_2022_CVPR}   &     & \textbf{93.4}  & 88.3  & 72.1 & 79.1 & -     & 89.0  \\ \midrule
\textbf{Ours}      &     &\underline{93.3}      & \textbf{89.4}      & \underline{73.3}     &    \textbf{80.5}  & \underline{80.4}      & \underline{89.4}      \\ 
\bottomrule
\end{tabular}
}
\vspace{-8mm}
\end{center}
\end{table}

\textbf{Results on CAMERA25 Dataset:}
The performance comparison of the proposed method and the state-of-the-art is shown in Table~\ref{tbl:cat_perform_camera}. Our method ranks top and second on all the metrics without prior information. Of the four scores ranked second, three are close to the tops with negligible differences ($0.1\%$ on $\tenDfiveC$ and $\IoUSevenFive$ metrics, and $0.2\%$ on $\fiveDtwoC$ metric). It is also worth noting that CAMERA25 is a synthetic dataset that contains no noise, so one main contribution of the proposed method, noise robustness, is not reflected in this dataset. However, this contribution can be identified by comparing the proposed and the state-of-the-art methods' performance on the CAMERA25 and the REAL275 dataset. The REAL275 dataset contains the same object categories as the CAMERA25 but is real-world collected and contains complex noise. It can be observed that the performance drop of our method is much less than other methods when encountering real-world noises in the REAL275. This demonstrates that our method is more noise-robust compared with other methods. A more comprehensive comparison with methods using RGB and RGB-D data is included in the supplementary, in which our method still shows competitive results despite using depth-only data.

\section{Conclusion}
In this paper, we proposed a hybrid scope latent feature extraction layer, the HS-layer, and used it to construct a category-level object pose estimation framework HS-Pose. Based on the advantages of the HS-layer, HS-Pose can handle complex shapes, capture an object's size and translation, and is robust to noise. The capability of the overall framework is demonstrated in the experiments. The comparisons with the existing methods show that our HS-Pose achieves state-of-the-art performance. In future work, we plan to apply our proposed HS-layer to other problems where unstructured data needs to be processed, and the combination between the local and the global information becomes critical. 




\section*{Acknowledgements}
This work was supported in part by the Institute of Information and communications Technology Planning and evaluation (IITP) grant funded by the Korea government (MSIT) (2021-0-00537), Benchmarks for UndeRstanding Grasping (BURG) (EP/S032487/1), National Natural Science Foundation of China under Grant No. 62073159 and Grant No. 62003155, Shenzhen Science and Technology Program under Grant No. JCYJ20200109141601708, and the Science, Technology and Innovation Commission of Shenzhen Municipality under grant no. ZDSYS20200811143601004.
\newpage
{\small
\bibliographystyle{ieee_fullname}
\bibliography{egbib}
}





\end{document}


\title{Supplementary Material of HS-Pose: Hybrid Scope Feature Extraction for Category-level Object Pose Estimation}

\author{
Linfang Zheng$^{1,4}$ \and Chen Wang$^{1,2}$ \and Yinghan Sun$^{1}$ \and Esha Dasgupta$^{4}$ \and Hua Chen$^{1}$ \and Ale\v{s} Leonardis$^{4}$ \and Wei Zhang\thanks{The corresponding author.} $^{1,3}$ \and Hyung Jin Chang$^{4}$ \and\\$^{1}$Department of Mechanical and Energy Engineering, Southern University of Science and Technology\\
$^{2}$Department of Computer Science, the University of Hong Kong\\
$^{3}$Peng Cheng Laboratory, Shenzhen, China\\
$^{4}$School of Computer Science, University of Birmingham\\
{\tt\small$\{$lxz948,exd949$\}$@student.bham.ac.uk, cwang5@cs.hku.hk, sunyh2021@mail.sustech.edu.cn}\\
{\tt\small $\{$chenh6,zhangw3$\}$@sustech.edu.cn,$\{$a.leonadis,h.j.chang$\}$@bham.ac.uk}
}

\maketitle

\section{Choice of the Baseline Method.}
We choose GPV-Pose~\cite{GPV-Pose_2022_CVPR} as the baseline for the following reasons:
\begin{itemize}
    \vspace{-1mm}
    \item \textbf{To demonstrate the effectiveness of the components in the HS-layer regarding pose estimation.} GPV-Pose is one of the state-of-the-art \textbf{3D-GC~\cite{3dgcn_lin_CVPR_2020} based} category-level object pose estimation methods. It is suitable for us to show how each component of the HS-layer incrementally added onto the 3D-GC layer influences the performance of pose estimation.
    \vspace{-2mm}
    \item \textbf{To compare the HS-layer with the strategies proposed by other methods.} For example, SSP-Pose~\cite{SSP-Pose_IROS_22} and RBP-Pose \cite{RBP-Pose_ECCV_2022} are also developed based upon GPV-Pose. The former leverages the prior-shape information and uses a shape deformation module to improve performance. The latter enhances GPV-Pose by a \textit{residual bounding box projection} (SPRV) module and a shape deformation module. We compare with SSP-Pose to demonstrate the effectiveness of STE. We also show the influence of the RF-F approach by comparing it with RBP-Pose. In the experiments, our simple STE and RF-F method outperform their counterparts in strict metrics (\eg, $\IoUSevenFive$, $\fiveDtwoC$, and $\fiveDfiveC$ metrics) and achieve competitive results in other metrics.
    \vspace{-2mm}
\end{itemize}

\section{About the Object Detector}
For a fair comparison, as when compared against other methods~\cite{GPV-Pose_2022_CVPR, RBP-Pose_ECCV_2022, SSP-Pose_IROS_22}, we also utilize the MaskRCNN~\cite{Maskrcnn_2017} to detect the objects in our experiments. It is worth noting that our method is not limited to MaskRCNN~\cite{Maskrcnn_2017}. Other object detectors such as SD-MaskRCNN~\cite{SD-MaskRCNN_ICRA_2019} and PointNet~\cite{Pointnet_2017_CVPR} can also be used. 

\section{About the Speed} 
Since the speed can be different when performed on different machines, we only use the results of the speed to demonstrate that our method can achieve real-time performance and do not emphasise a speed comparison with other methods. 

\subsection{The Speed of GPV-Pose}
For a fair speed comparison with the baseline, GPV-Pose~\cite{GPV-Pose_2022_CVPR}, we report the speed of GPV-Pose on our machine with the same evaluation code as ours. The speed of GPV-Pose achieved on our machine (69 FPS) is faster than the original paper (20 FPS) due to the following reasons:
\begin{itemize}
    \vspace{-1mm}
    \item  \textbf{The difference between the machines.} The original paper of GPV-Pose reports the speed test on a single TITAN X GPU, while we test GPV-Pose on a single RTX 3090 GPU with an Intel(R) Core(TM) i9-10900K CPU, 32 GB RAM. The speed is 33 FPS on our machine.
    \vspace{-2mm}
    \item \textbf{The difference in the evaluation code.} Our evaluation code is a refactored version of GPV-Pose's code. We change some for-loop operations to batch operations and remove unnecessary calculations (\eg the bounding box voting and symmetric point cloud reconstruction) during inference. These changes significantly boost the speed from 33 FPS to 69 FPS. All the changes have passed unit tests to ensure they get the same results as the original code.  
\end{itemize}

\begin{table*}[!h]
\begin{center}

\caption{\textbf{Comparison with the state-of-the-art methods on REAL275 dataset.} Overall best results are in bold, and the second-best results are underlined. \textit{Type} lists the type of input data for
pose estimation. \textit{Syn.} denotes whether the synthetic data is used during training.}
\label{tbl:cat_perform_all}
\vspace{-3mm}
\resizebox{0.98\linewidth}{!}
{%
\footnotesize

\begin{tabular}{@{}r|c|c|ccc|ccccc|c@{}}
\toprule

Method & Type & Syn. & $\text{IoU}_{25}$   &$\text{IoU}_{50}$    &$\text{IoU}_{75}$     &$5^\circ 2$cm &$5^\circ 5$cm &$10^\circ 2$cm &$10^\circ 5$cm &$10^\circ 10$cm&Speed(FPS)  \\
\midrule            
\midrule            
NOCS~\cite{NOCS_2019_CVPR}&RGB-D& \checkmark&\textbf{84.9}&80.5&30.1& -& 9.5        &13.8 &26.7&26.7&5\\           
CASS~\cite{CASS_2020_CVPR}&RGB-D & \checkmark &{84.2} &77.7 &15.3 &19.5 &23.5 &50.8 &58.0 &58.3 & -\\
SPD~\cite{Tian_ShapePrior_2020_ECCV}  & RGB-D & \checkmark &{83.4} &77.3 &53.2 & 19.3 & 21.4 &43.2 &54.1 &-  & 4 \\
DualPoseNet~\cite{DualPoseNet_2021_ICCV}&RGB-D & \checkmark &- &79.8 &62.2 &29.3 &35.9 &50.0 &66.8 &- &2\\
SGPA~\cite{SGPA_2021_ICCV} &RGB-D & \checkmark &- &80.1 &61.9 &{35.9} &39.6 &{61.3} &70.7 &- &-\\
CR-Net~\cite{CR-Net_2021_IROS} &RGB-D & \checkmark &- &79.3 &55.9 &27.8 &34.3 &47.2 &60.8&- &-\\
Self-DPDN \cite{Self-DPDN_ECCV_2022} & RGB-D & \checkmark & - &\textbf{83.4} & \textbf{76.0} &{46.0} &{50.7} &\textbf{70.4} &78.4 &-& -\\

\midrule
SPD~\cite{Tian_ShapePrior_2020_ECCV}   &RGB & \checkmark &- &75.2 &46.5 &15.7 &18.8 &33.7  &47.4 &-&4\\
\midrule
SAR-Net \cite{SAR-Net_Lin_2022_CVPR} &D & &- &79.3 &62.4 &31.6 &{42.3} &50.4 &68.3 &- &10\\

FS-Net\tablefootnote{We use the results provided by the GPV-Pose, which uses the GPV-Pose's decoder for a fair comparison and shows higher performance than the originally reported results of the FS-Net.} \cite{FSNET_2021_CVPR} &D & &84.0 &81.1 &63.5 &19.9 &33.9 &- &69.1 &71.0 &20\\


SSP-Pose \cite{SSP-Pose_IROS_22} & D & &84.0 &82.3 &66.3 &34.7 &44.6 & - &77.8 &{79.7} & {25}\\ 

RBP-Pose\cite{RBP-Pose_ECCV_2022} & D & & - & - &{67.8} &38.2 &48.1 & 63.1 &{79.2} &- & 25 \\

GPV-Pose \cite{GPV-Pose_2022_CVPR} &D & &{84.1} &\underline{83.0}&{64.4}&32.0 &{42.9} &55.0 &{73.3}  &{74.6}&\textbf{69}\\

\midrule

\textbf{Ours (10 neighbors)} &D & &{84.2} & {82.1} &{74.7} &\textbf{46.5} &\underline{55.2} &{68.6} &\underline{82.7} &\underline{83.7} &\underline{50}\\ 

\textbf{Ours (20 neighbors)} &D & &\underline{84.3} & {82.8} &\underline{75.3} &\underline{46.2} &\textbf{56.1} &\underline{68.9} &\textbf{84.1} &\textbf{85.2} &38\\ 

\bottomrule
\end{tabular}
}
\vspace{-15px}
\end{center}
\end{table*}
\begin{figure}[ht]
\centering
\includegraphics[width=1\linewidth, trim = 220 0 260 20, clip]{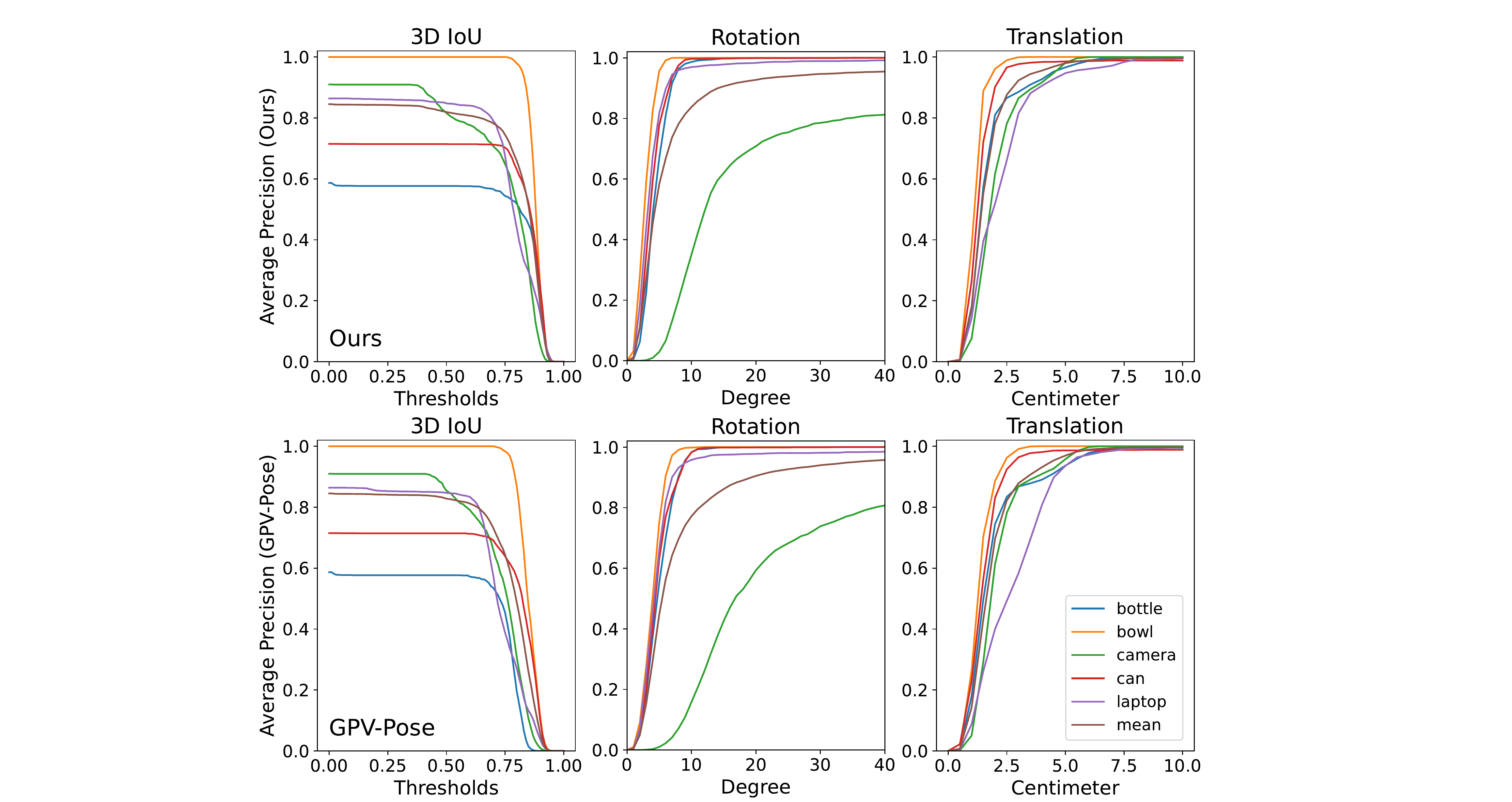}
\vspace{-7mm}
\caption{\footnotesize \textbf{Per-category comparison between our method and GPV-Pose.} We demonstrate average precision \emph{v.s.} different error thresholds on the REAL275 dataset.}

\vspace{-6mm}
\label{fig:mAP_cirve}
\end{figure} 
\section{Comparison with State-of-the-Arts Methods}
We add the comparison between the proposed HS-Pose with methods that use different data modalities (\eg RGB and RGB-D) in this section.

\subsection{Results on REAL275 dataset.} 
The comparison between the proposed method and the state-of-the-art methods on the REAL275 dataset is shown in Table~\ref{tbl:cat_perform_all}. Our method outperforms the depth-only methods in 7 out of 8 pose estimation and size estimation metrics and achieves comparable performance in the remaining metric. Our depth-only method also achieves competitive results with the RGB-D-based approaches and outperforms them in several pose estimation metrics (\eg $\textbf{56.1\%}$ (ours) vs. $50.7\%$ on $\fiveDfiveC$ metric). It is worth noting that many of them are trained with synthetic data or using CAMERA25 and REAL275 for mixed training, which results in a large number of training images and many more objects (over 1K objects for CAMERA25 and Real275 mixed training) for training. In contrast, our method is trained on 1.6K real images of 18 objects. In Figure~\ref{fig:mAP_cirve}, we present the average precision of each category under different thresholds and compare it with the GPV-Pose.

\begin{table}[!t]
\begin{center}

\caption{\textbf{Comparison with state-of-the-art methods on CAMERA25 dataset.} Overall best results are in bold, and the second-best results are underlined. \emph{Type} lists the type of input data for pose estimation. \emph{Prior} denotes whether the method uses shape priors.}
\label{tbl:cat_perform_camera_full}
\vspace{-3mm}
\resizebox{\linewidth}{!}
{%

\begin{tabular}{@{}r|c|c|cc|cccc@{}}
\toprule

Method & Type & Prior & $\text{IoU}_{50}$ & $\text{IoU}_{75}$ & $5^\circ 2\text{cm}$ & $5^\circ 5\text{cm}$ & $10^\circ 2\text{cm}$ & $10^\circ 5\text{cm}$    \\ 
\midrule
\midrule            

SPD\cite{Tian_ShapePrior_2020_ECCV} & RGB-D   & \checkmark   & 93.2  & 83.1  & 54.3 & 59.0 & 73.3  & 81.5  \\
CR-Net \cite{CR-Net_2021_IROS}& RGB-D & \checkmark   & \textbf{93.8}  & 88.0  & 72.0 & 76.4 & 81.0  & 87.7  \\
SGPA \cite{SGPA_2021_ICCV}  & RGB-D    & \checkmark   & 93.2  & 88.1  & 70.7 & 74.5 & \textbf{82.7}  & 88.4  \\
NOCS \cite{NOCS_2019_CVPR}  &  RGB-D   &     & 83.9  & 69.5  & 32.3 & 40.9 & 48.2  & 64.6  \\
DualPoseNet \cite{DualPoseNet_2021_ICCV} & RGB-D  &     & 92.4  & 86.4  & 64.7 & 70.7 & 77.2  & 84.7  \\
\midrule            
SPD\cite{Tian_ShapePrior_2020_ECCV} & RGB   & \checkmark   & 93.1  & 84.6  & 50.2 & 54.5 & 70.4  & 78.6  \\
\midrule            
SAR-Net\cite{SAR-Net_Lin_2022_CVPR} &  D  &\checkmark  & 86.8  & 79.0  & 66.7 & 70.9 & 75.3  & 80.3  \\
SSP-Pose\cite{SSP-Pose_IROS_22} &  D  & \checkmark  & -  & 86.8  & 64.7 & 75.5 & -  & 87.4  \\
RBP-Pose \cite{RBP-Pose_ECCV_2022}      & D  & \checkmark   & 93.1  & {89.0}  & \underline{73.5} & {79.6} & \underline{82.1}  & \underline{89.5}  \\ 
GPV-Pose \cite{GPV-Pose_2022_CVPR} & D  &     & \underline{93.4}  & 88.3  & 72.1 & 79.1 & -     & 89.0  \\ \midrule

\textbf{Ours (10 neighbors)}   &  D   &     &{93.3}      & \textbf{89.4}      & {73.3}     &    \underline{80.5}  & {80.4}      & {89.4}      \\ 
\textbf{Ours (20 neighbors)}   &  D   &     &\underline{93.4}      & \underline{89.3}      & \textbf{74.0}     & \textbf{82.0}  & {80.3}      & \textbf{90.2}      \\ 
\bottomrule
\end{tabular}
}
\vspace{-15px}
\end{center}
\end{table}

\begin{table*}[!ht]
\begin{center}

\caption{\textbf{Per-category results of our method on REAL275 dataset.} }
\label{tbl:per_cat_real275}
\vspace{-3mm}
\resizebox{0.95\linewidth}{!}
{%

\begin{tabular}{@{}c|ccc|ccccc|cc|ccc@{}}
\toprule

category & $\IoUTwoFive$ & $\IoUFifty$ & $\IoUSevenFive$ & $\fiveDtwoC$ & $\fiveDfiveC$ & $\tenDtwoC$ & $\tenDfiveC$ & $10^\circ 10\text{cm}$ & $5^\circ $   & $10^\circ $   & $2$cm   & $5$cm    & $10$cm   \\ 
\midrule
\midrule            
bottle   & 57.7  & 57.7  & 54.8  & 43.0 & 53.1 & 80.0  & 95.4  & 98.5   & 66.9 & 99.2  & 81.0 & 96.5  & 99.5  \\
bowl     & 100.0 & 100.0 & 100.0 & 92.1 & 95.6 & 96.5  & 100.0 & 100.0  & 95.6 & 100.0 & 96.5 & 100.0 & 100.0 \\
camera   & 90.9  & 82.3  & 65.2  & 2.3  & 3.1  & 28.3  & 35.7  & 35.8   & 3.1  & 35.8  & 60.9 & 98.1  & 100.0 \\
can      & 71.4  & 71.4  & 70.5  & 68.6 & 75.0 & 90.0  & 98.5  & 98.5   & 77.8 & 99.7  & 90.0 & 98.6  & 98.8  \\
laptop   & 86.1  & 84.9  & 67.1  & 49.3 & 79.5 & 52.4  & 94.0  & 96.8   & 80.8 & 96.9  & 52.4 & 94.6  & 99.5  \\
mug      & 99.2  & 96.4  & 90.8 & 23.8 & 25.0 & 64.6  & 72.6  & 72.6   & 25.0 & 72.6  & 88.5 & 100.0 & 100.0 \\ \midrule
average  & 84.2  & 82.1  & 74.7  & 46.5 & 55.2 & 68.6  & 82.7  & 83.7   & 58.2 & 84.0  & 78.2 & 98.0  & 99.6  \\ \bottomrule
\end{tabular}
}
\vspace{-10px}
\end{center}
\end{table*}
\begin{table*}[ht]
\begin{center}

\caption{\textbf{Per-category results of our method on CAMERA25 dataset.} }
\label{tbl:per_cat_camera25}
\vspace{-3mm}
\resizebox{0.95\linewidth}{!}
{%

\begin{tabular}{@{}c|ccc|ccccc|cc|ccc@{}}
\toprule

category & $\text{IoU}_{25}$ & $\text{IoU}_{50}$ & $\text{IoU}_{75}$ & $5^\circ 2\text{cm}$ & $5^\circ 5\text{cm}$ & $10^\circ 2\text{cm}$ & $10^\circ 5\text{cm}$ & $10^\circ 10\text{cm}$ & $5^\circ $   & $10^\circ $   & $2$cm   & $5$cm    & $10$cm   \\ 
\midrule
\midrule            
bottle  &93.9  &93.8  &90.9  &80.1  &96.7  &80.7  &97.8  &99.4  &98.5  &99.8  &80.7  &97.9  &99.5\\
bowl  &96.9  &96.8  &96.8  &98.4  &98.6  &99.4  &99.8  &99.8  &98.7  &99.8  &99.4  &99.8  &99.9\\
camera  &94.8  &85.4  &74.3  &51.2  &55.1  &65.0  &70.6  &70.9  &55.5  &71.4  &86.9  &99.0  &99.6\\
can  &92.5  &92.4  &92.2  &99.0  &99.4  &99.0  &99.5  &99.5  &99.9  &100.0  &99.0  &99.5  &99.6\\
laptop  &98.4  &97.4  &90.6  &75.6  &85.2  &81.1  &92.7  &97.0  &89.0  &97.1  &83.3  &95.6  &99.9\\
mug  &94.1  &93.8  &91.9  &35.4  &47.9  &57.4  &76.2  &76.2  &49.1  &76.9  &75.9  &99.5  &99.6\\ \midrule
average  &95.1  &93.3  &89.4  &73.3  &80.5  &80.4  &89.4  &90.5  &81.8  &90.8  &87.5  &98.6  &99.7\\ \bottomrule
\end{tabular}
}
\vspace{-15px}
\end{center}
\end{table*}

\subsection{Results on CAMERA25 dataset.}
We test the proposed method on the CAMERA25 dataset and show the comparison results of the proposed method with other approaches in Table~\ref{tbl:cat_perform_camera_full}. We achieved top and second scores on 5 out of 6 metrics (4 tops and 1 second) with no need for RGB data. 

\section{Per-category Results}
The per-category results trained on the REAL275 and CAMERA25 datasets are shown in Table~\ref{tbl:per_cat_real275} and Table~\ref{tbl:per_cat_camera25}, respectively. 

\section{Settings of Noise Resistance Experiments}
In the ablation study [AS-6], we compared the outlier robustness of the proposed method and the baseline. We define \textit{outliers} as the points that do not belong to the target object. The \textit{outlier ratio} is defined as the ratio of the outliers' number to the total point number of the input point cloud. We use the REAL275 dataset for testing and generate the noisy input data by sampling points from the background and the object region according to the outlier ratio.
To ensure a fair comparison, the noisy data used for testing the proposed and baseline methods is the same. 

\begin{figure}[]
\centering
\vspace{-4mm}
\includegraphics[width=0.75\linewidth, trim = 0 20 20 20, clip]{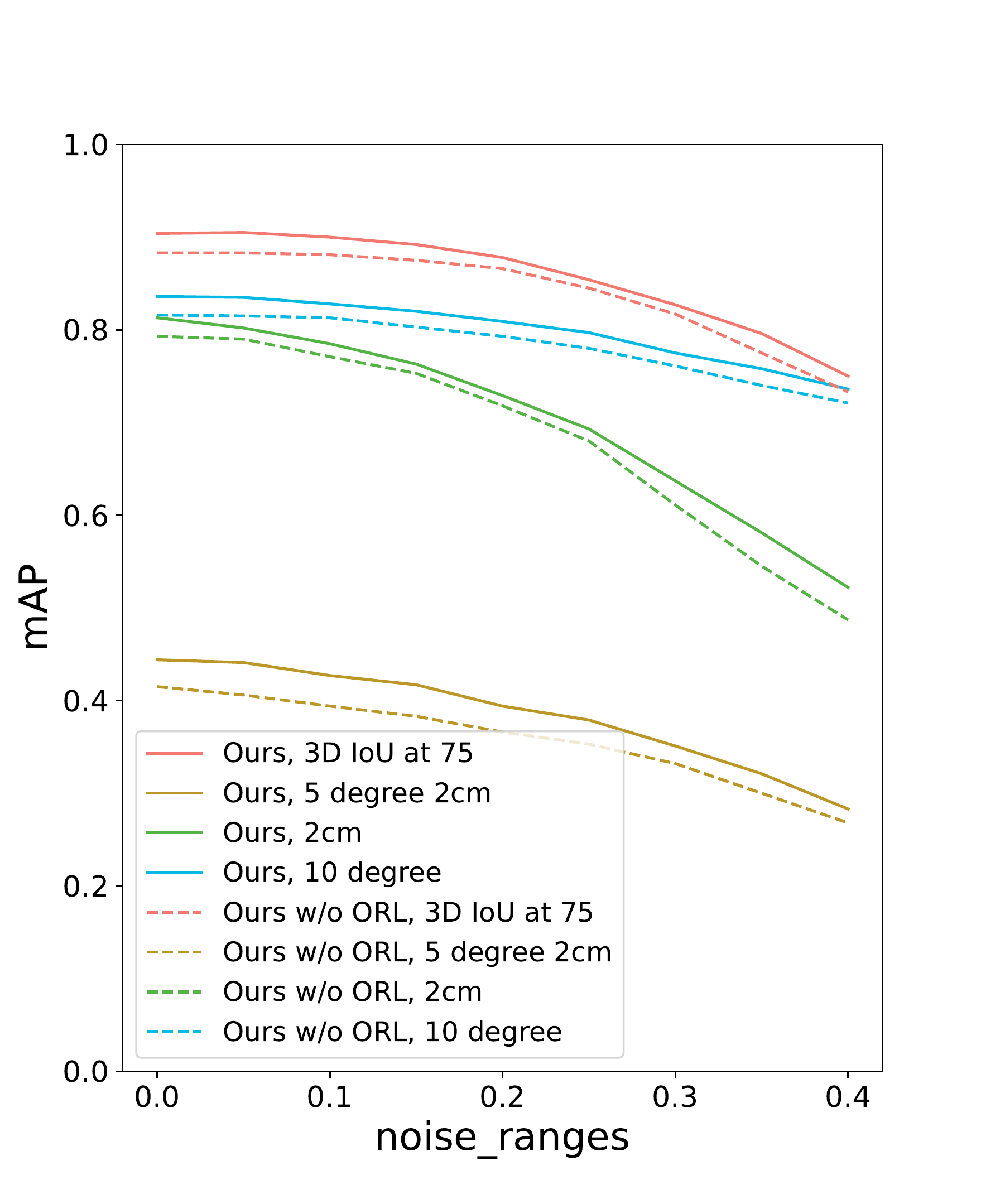}
\vspace{-4mm}
\caption{\footnotesize \textbf{The comparison of noise resistance of the proposed HS-Pose with (Ours) and without outlier robust feature extraction layer (ORL) (Ours w/o ORL).} We show the influences of ORL in differences metrics (\eg $2\text{cm}$, $\text{IoU}_{75}$, $5^\circ2\text{cm}$, and $10^\circ$). After adding ORL, the performance is enhanced across different noise levels (from 0.0\% to 40.0\% outliers). 
} 
\vspace{-15px}
\label{fig:noise_resistance_ORL}
\end{figure} 

\section{Ablation Study on ORL}
We demonstrate the effectiveness of the proposed outlier robust feature extraction layer (ORL) in Fig.~\ref{fig:noise_resistance_ORL}. We test the noise resistance of the proposed HS-Pose with and without ORL using different outlier ratios (from 0.0\% to 40.0\%) in the input point cloud. The figure shows that the ORL successfully enhances the performance on the size-pose-joint metric ($\text{IoU}_{75}$), translation metric ($2\text{cm}$), and rotation metric ($10^\circ$) across different noise levels.

\section{Ablation Study on the Neighbor Numbers}
To investigate the influences of the neighbor numbers, we test the performance of the proposed method using different neighbor numbers (from 3 to 40) in the RF-F and ORL\footnote{Due to the limit of the GPU memory size, we set the batch size to 16, 16, and 8 for 20, 30, and 40 neighbors, respectively.}. The experiments are separated into three groups to evaluate the impact: 1) change the RF-F's neighbor number with the ORL's neighbor number fixed, 2) change the ORL's neighbor number with the RF-F's neighbor number fixed, and 3) change the neighbor numbers of the RF-F and ORL simultaneously. 


\subsection{Change RF-F's Neighbor Number Only}
\begin{table}
\begin{center}

\caption{\textbf{Performance of the proposed method when changing the neighbor number of RF-F.} The neighbor number of the ORL is fixed to 10 in this experiment. Overall best results are in bold, and the second-best results are underlined.}
\label{tbl:RFF_neighbor}
\vspace{-3mm}
\resizebox{1\linewidth}{!}
{\footnotesize

\begin{tabular}{@{}c|cccccc@{}}
\toprule

Neighbor Number & 3 &5   &10    &20    &30 & 40  \\
\midrule            
\midrule   
$5^\circ 2\text{cm}$ &39.8    &41.5   &\textbf{46.5}   &\underline{46.1}   &44.3    &41.6   \\ 
$5^\circ 5\text{cm}$ &49.2    &51.4   &\underline{55.2}   &\textbf{56.7}   &54.7    &54.4    \\ 
$\text{IoU}_{75}$    &72.9    &72.8   &\textbf{74.7}   &\underline{73.4}   &\textbf{74.7}    &71.9    \\ 
Speed (FPS)          &\textbf{64}      &60     &50     &41     &34      &30 \\








\bottomrule
\end{tabular}
}
\vspace{-3mm}
\end{center}
\end{table}
Table~\ref{tbl:RFF_neighbor} shows the performance of the proposed method using different neighbor numbers in the RF-F with the ORL's neighbor number fixed to 10. As seen from the table, finding more neighboring 3D points using feature distance requires a longer time, while a certain range of neighbor numbers (around 10-20 neighbors) produces better precision than other numbers. Specifically, the speed decreased from 64 FPS to 30 FPS when increasing the neighbor number from 3 to 40. In the meantime, the performance on $5^\circ 2\text{cm}$, which starts at $39.8\%$, reaches its best at $46.5\%$ when using 10 neighbors, after which it begins to decline and ultimately reaches a score of $41.6\%$ at 40 neighbors. Generally, using 10 neighbors for RF-F achieves the overall best performance while maintaining fast speed. The reason why insufficient and excessive neighbor numbers adversely affect the precision might be that fewer neighbors cannot fully characterize the global geometric feature, whereas an excessive number of neighbors may obscure the geometric structural information in the formed receptive field.

\subsection{Change ORL's Neighbor Number Only}
\begin{table}
\begin{center}

\caption{\textbf{Performance of the proposed method when changing the neighbor number of ORL.} The neighbor number of RF-F is fixed to 10 in this experiment. Overall best results are in bold, and the second-best results are underlined.}
\label{tbl:ORL_neighbor}
\vspace{-3mm}
\resizebox{1\linewidth}{!}
{\footnotesize

\begin{tabular}{@{}c|cccccc@{}}
\toprule

Neighbor Number & 3 &5   &10    &20    &30 & 40  \\
\midrule            
\midrule   
$5^\circ 2\text{cm}$ &43.3    &\underline{43.6}   &\textbf{46.5}&42.7   &43.1    &39.4   \\ 
$5^\circ 5\text{cm}$ &53.1    &53.0   &\underline{55.2}   &\textbf{55.3}   &54.4    &53.7    \\ 
$\text{IoU}_{75}$    &74.6   &72.8   &\textbf{74.7}   &72.7   &73.9    &71.1    \\ 
Speed (FPS)          &\textbf{52}      &51     &50     &48     &48      &49 \\

\bottomrule
\end{tabular}
}
\vspace{-7mm}
\end{center}
\end{table}
Table~\ref{tbl:ORL_neighbor} shows the performance of the proposed method using different neighbor numbers in the ORL with the RF-F's neighbor number fixed to 10. According to the table, the speed for finding neighboring points in 3D space is relatively stable, which only dropped by 4 FPS when the neighbor number increased from 3 to 40. Compared to the RF-F, the neighbor number impacts the speed less. The reason is that in RF-F, the nearest neighbors are found in higher dimensional feature space. In terms of precision, an appropriate range of neighbor numbers is beneficial for ORL to balance finding the reliable points and outliers. Similar to RF-R, using 10 neighbors performs better than other values in our experiments.

\subsection{Change the Neighbor Number Simultaneously}
\begin{table}
\begin{center}
\caption{\textbf{Performance of the proposed method when changing the neighbor number of the ORL and RF-F together.} The neighbor number of ORL and RF-F are set to the same in this experiment. Overall best results are in bold, and the second-best results are underlined.}
\label{tbl:ORL_RFF_neighbor}
\vspace{-3mm}
\resizebox{1\linewidth}{!}
{\footnotesize

\begin{tabular}{@{}c|cccccc@{}}
\toprule

Neighbor Number & 3 &5   &10    &20    &30 & 40  \\
\midrule            
\midrule   
$5^\circ 2\text{cm}$ &39.0    &41.7   &\textbf{46.5}&{46.2}   &\underline{46.4}    &39.6   \\ 
$5^\circ 5\text{cm}$ &47.6    &52.8   &55.2   &\underline{56.1}   &\textbf{56.6}    &55.8    \\ 
$\text{IoU}_{75}$    &73.1    &72.7   &74.7   &\textbf{75.3}   &\underline{75.2}    &70.3    \\ 
Speed (FPS)          &64      &59      &50     &38     &30      &26 \\
\bottomrule
\end{tabular}
}
\vspace{-6mm}
\end{center}
\end{table}


 

Table~\ref{tbl:ORL_RFF_neighbor} shows the performance of the proposed method with the neighbor numbers of the ORL and RF-F changing simultaneously.
As shown in the table, when the neighbor numbers are around 10-30, the performance of our method is best. Moreover, in this range, using the same number of neighbors leads to better precision compared to fixing one of the neighbor numbers to 10. The reason might be that with increasing neighbor number in RF-F, more global geometric structure information can be found, and the possibility to include uninformed points is also increased. Therefore, with the neighbor number in ORL also increased, the effect brought by these uninformed points can be compensated, thus resulting a better performance. However, with too many neighbors, the performance still deteriorates because the balance between identifying reliable points and rejecting outliers is hurt.

\section{Qualitative Results}
\begin{figure*}[htbp]
	\centering
	\begin{minipage}{0.06\linewidth}
	    scene 1
	\end{minipage}
	\begin{minipage}{0.23\linewidth}
		\centering
		\includegraphics[width=\linewidth]{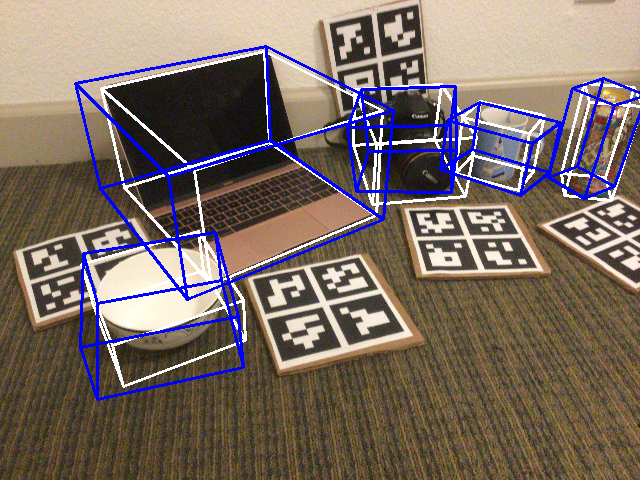}
	\end{minipage}
	\begin{minipage}{0.23\linewidth}
		\centering
		\includegraphics[width=\linewidth]{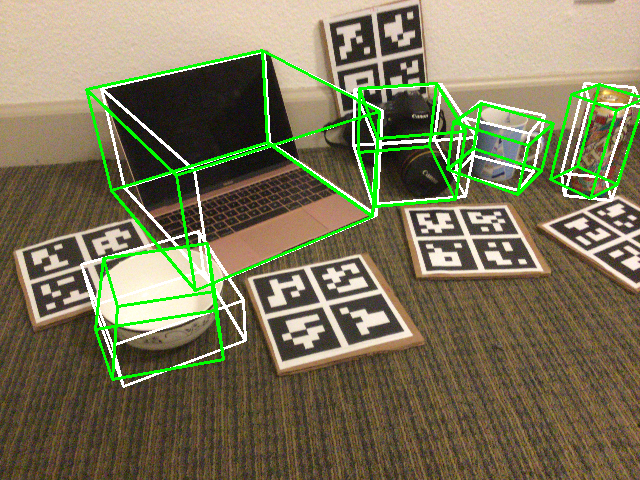}
	\end{minipage}
	\begin{minipage}{0.23\linewidth}
		\centering
		\includegraphics[width=\linewidth]{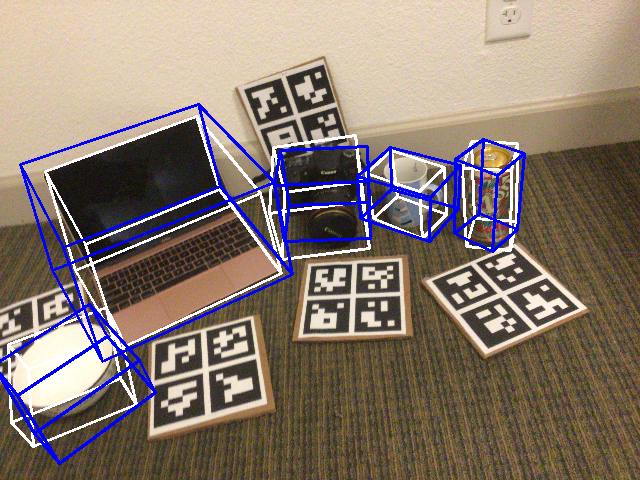}
	\end{minipage}
	\begin{minipage}{0.23\linewidth}
		\centering
		\includegraphics[width=\linewidth]{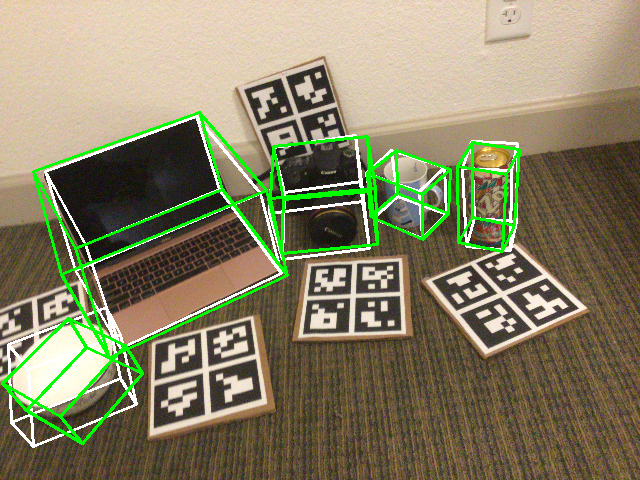}
	\end{minipage}
	
	\begin{minipage}{0.06\linewidth}
	    scene 2
	\end{minipage}
	\begin{minipage}{0.23\linewidth}
		\centering
		\includegraphics[width=\linewidth]{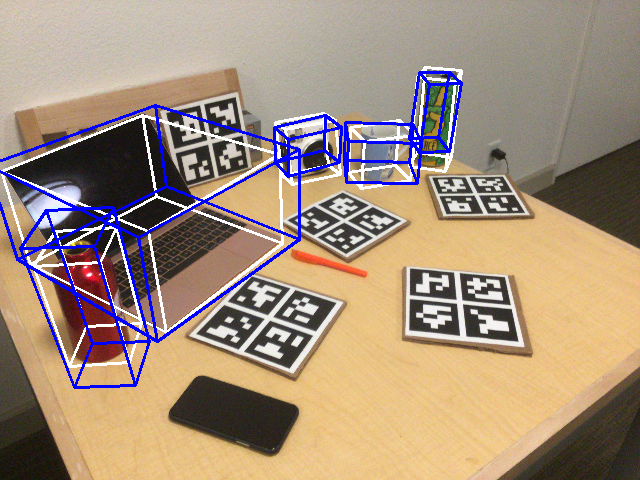}
	\end{minipage}
	\begin{minipage}{0.23\linewidth}
		\centering
		\includegraphics[width=\linewidth]{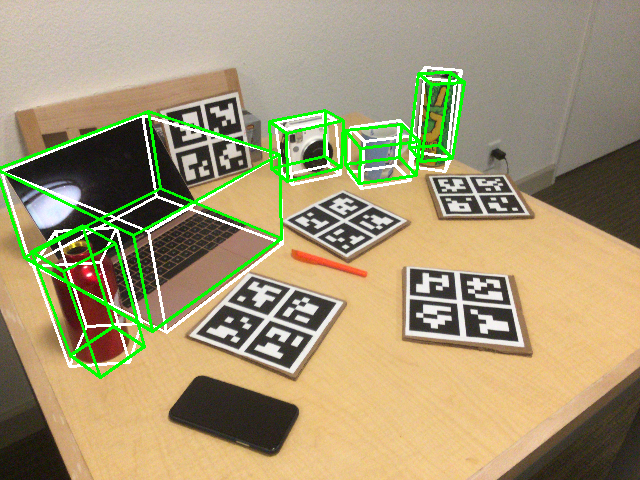}
	\end{minipage}
	\begin{minipage}{0.23\linewidth}
		\centering
		\includegraphics[width=\linewidth]{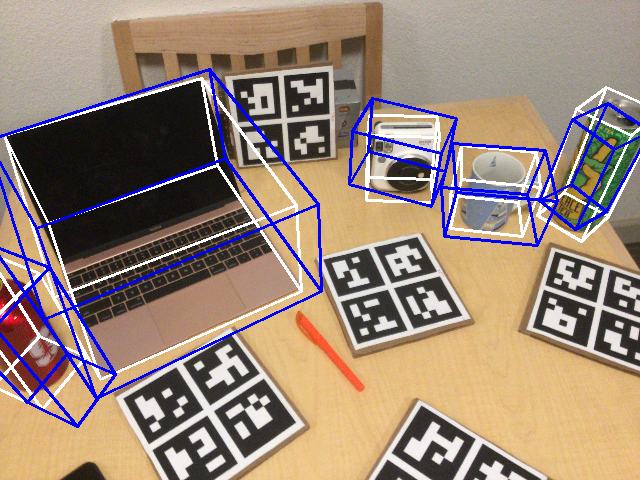}
	\end{minipage}
	\begin{minipage}{0.23\linewidth}
		\centering
		\includegraphics[width=\linewidth]{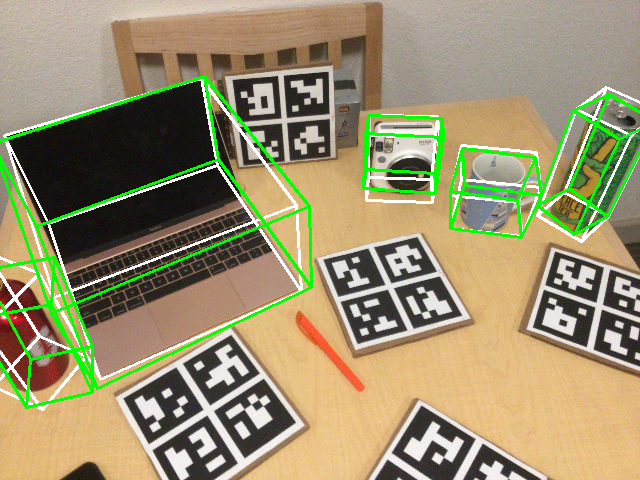}
	\end{minipage}
	
	\begin{minipage}{0.06\linewidth}
	    scene 3
	\end{minipage}
	\begin{minipage}{0.23\linewidth}
		\centering
		\includegraphics[width=\linewidth]{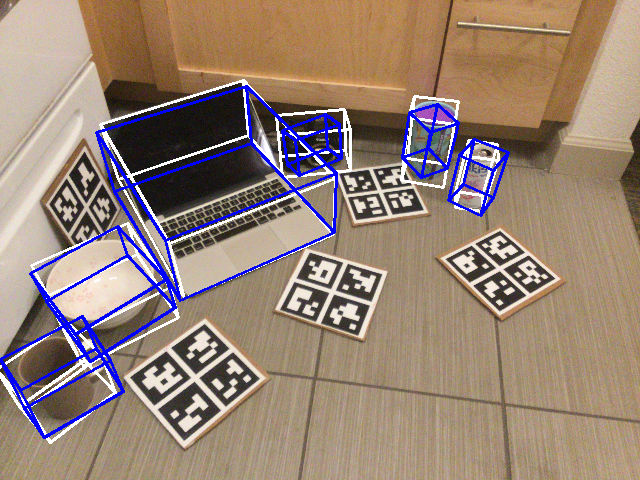}
	\end{minipage}
	\begin{minipage}{0.23\linewidth}
		\centering
		\includegraphics[width=\linewidth]{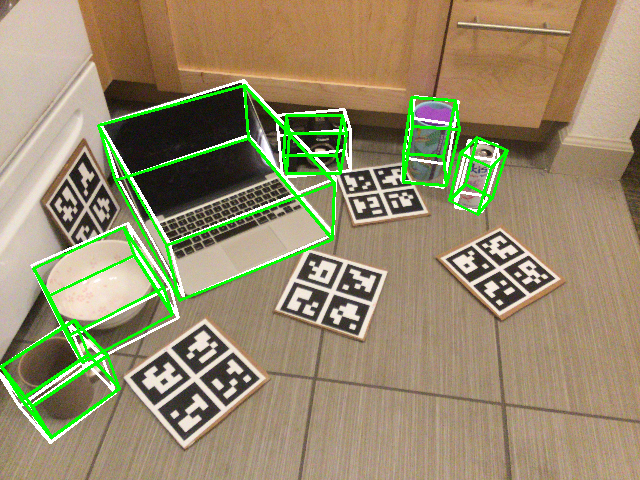}
	\end{minipage}
	\begin{minipage}{0.23\linewidth}
		\centering
		\includegraphics[width=\linewidth]{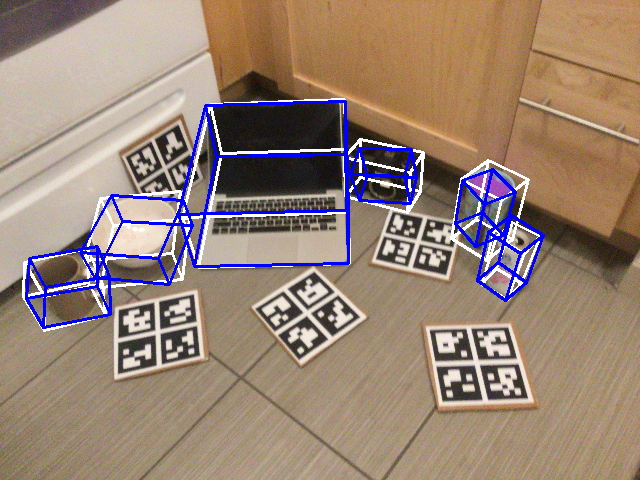}
	\end{minipage}
	\begin{minipage}{0.23\linewidth}
		\centering
		\includegraphics[width=\linewidth]{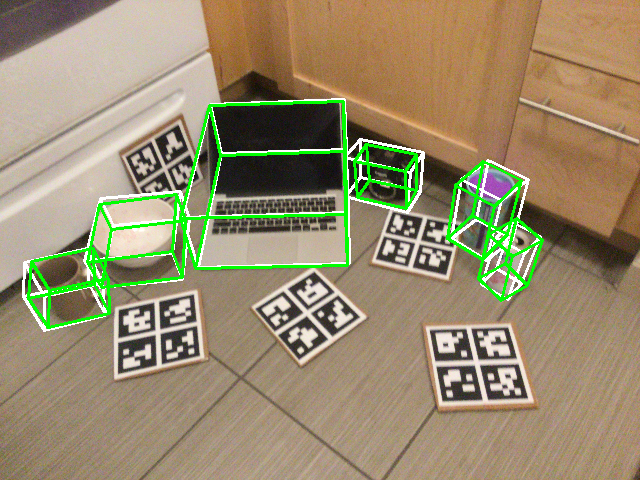}
	\end{minipage}
	
	\begin{minipage}{0.06\linewidth}
	    scene 4
	\end{minipage}
	\begin{minipage}{0.23\linewidth}
		\centering
		\includegraphics[width=\linewidth]{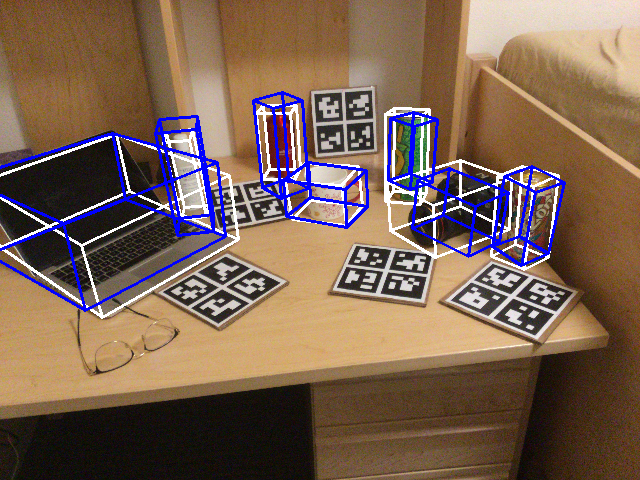}
	\end{minipage}
	\begin{minipage}{0.23\linewidth}
		\centering
		\includegraphics[width=\linewidth]{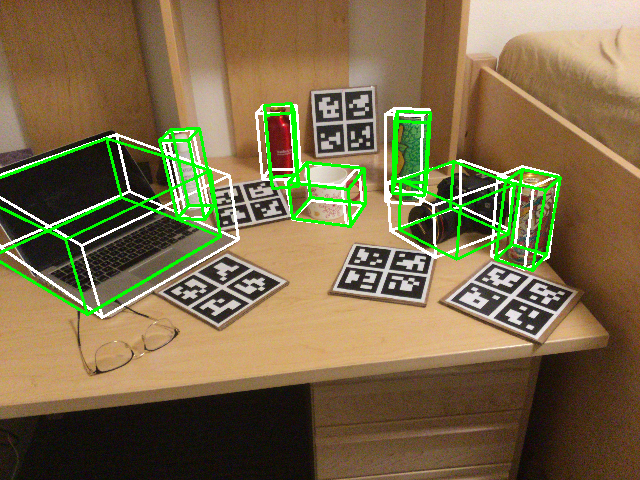}
	\end{minipage}
	\begin{minipage}{0.23\linewidth}
		\centering
		\includegraphics[width=\linewidth]{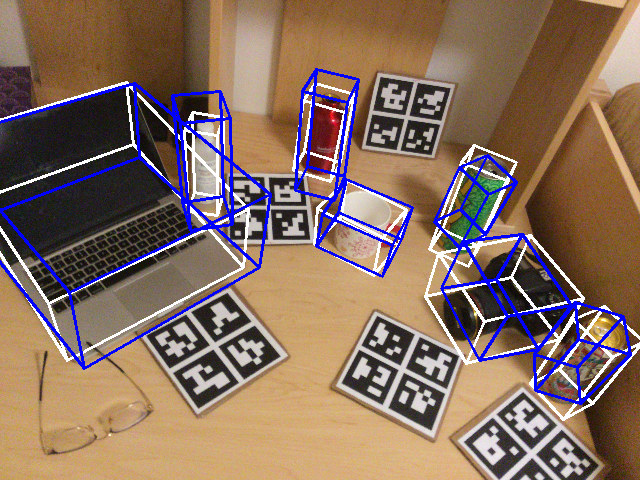}
	\end{minipage}
	\begin{minipage}{0.23\linewidth}
		\centering
		\includegraphics[width=\linewidth]{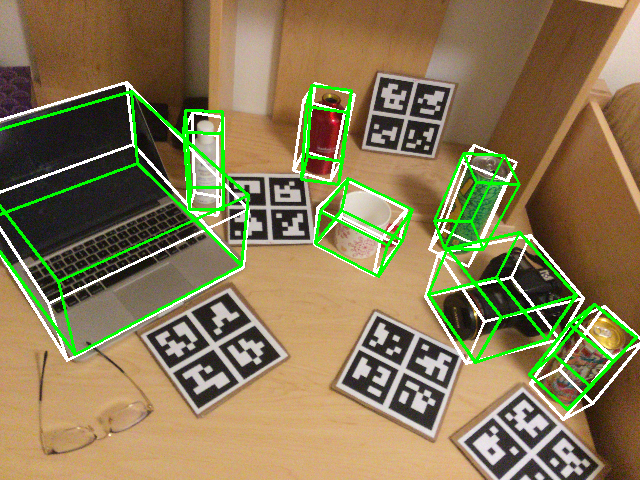}
	\end{minipage}
	
	\begin{minipage}{0.06\linewidth}
	    scene 5
	\end{minipage}
	\begin{minipage}{0.23\linewidth}
		\centering
		\includegraphics[width=\linewidth]{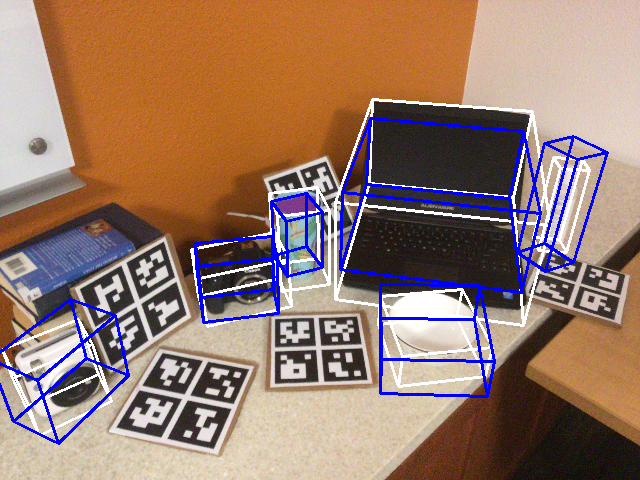}
	\end{minipage}
	\begin{minipage}{0.23\linewidth}
		\centering
		\includegraphics[width=\linewidth]{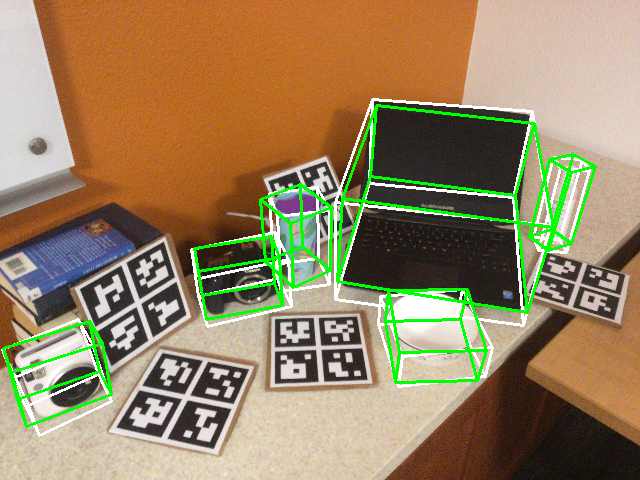}
	\end{minipage}
	\begin{minipage}{0.23\linewidth}
		\centering
		\includegraphics[width=\linewidth]{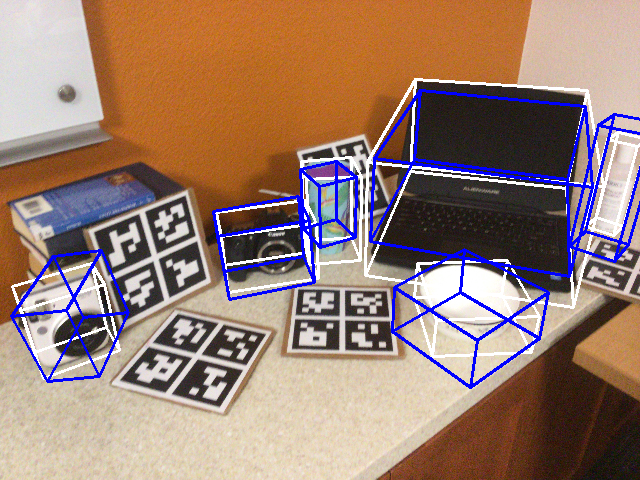}
	\end{minipage}
	\begin{minipage}{0.23\linewidth}
		\centering
		\includegraphics[width=\linewidth]{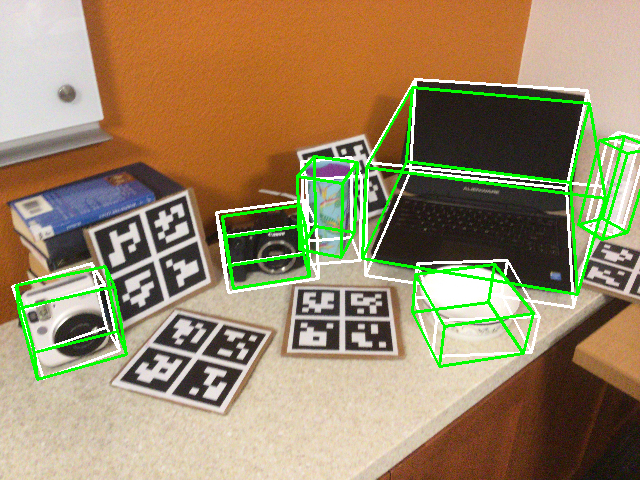}
	\end{minipage}
	
	\begin{minipage}{0.06\linewidth}
	    scene 6
	\end{minipage}
	\begin{minipage}{0.23\linewidth}
		\centering
		\includegraphics[width=\linewidth]{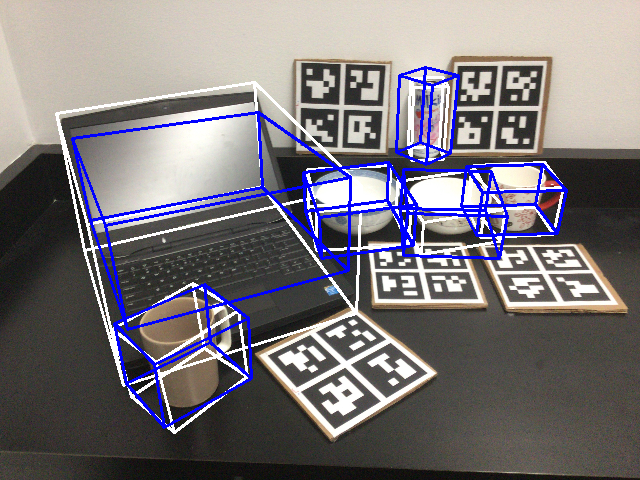}
	\end{minipage}
	\begin{minipage}{0.23\linewidth}
		\centering
		\includegraphics[width=\linewidth]{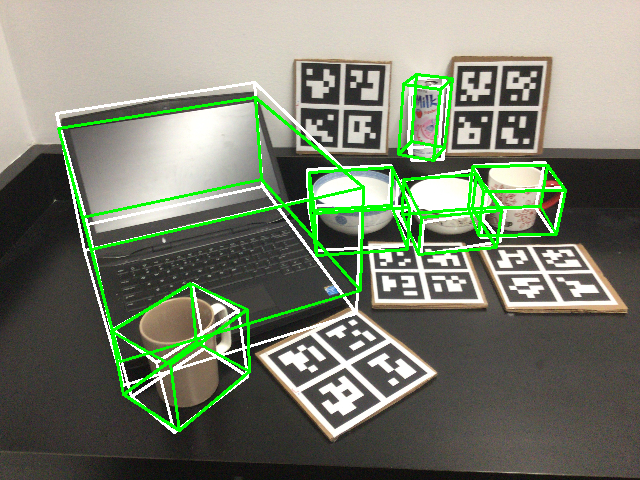}
	\end{minipage}
	\begin{minipage}{0.23\linewidth}
		\centering
		\includegraphics[width=\linewidth]{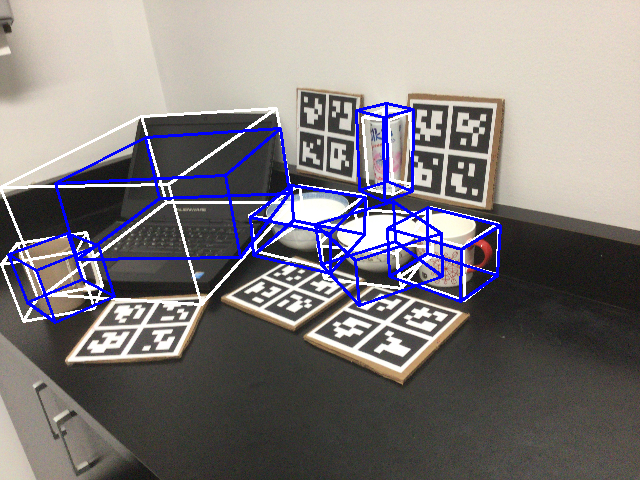}
	\end{minipage}
	\begin{minipage}{0.23\linewidth}
		\centering
		\includegraphics[width=\linewidth]{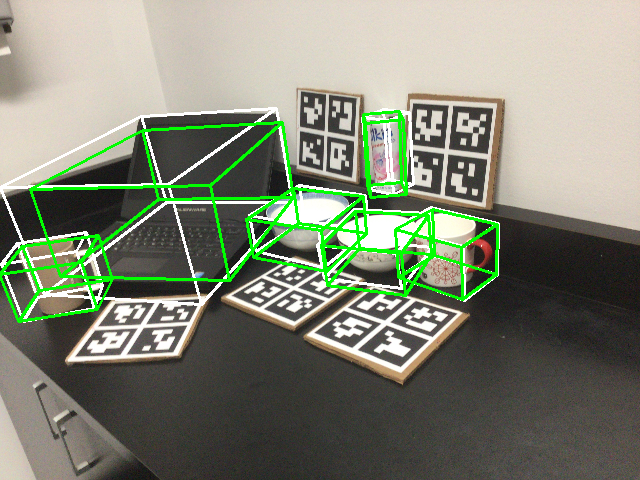}
	\end{minipage}
	
	
	\begin{minipage}{0.06\linewidth}
	{\color{white} scene 7}
	\end{minipage}
	\begin{minipage}[htbp]{0.23\linewidth}
	    \vspace{1mm}
	    \centering
		GPV-Pose
	\end{minipage}
	\begin{minipage}[htbp]{0.23\linewidth}
	    \vspace{1mm}
	    \centering
		Ours
	\end{minipage}
	\begin{minipage}[htbp]{0.23\linewidth}
	    \vspace{1mm}
	    \centering
		GPV-Pose
	\end{minipage}
	\begin{minipage}[htbp]{0.23\linewidth}
	    \vspace{1mm}
	    \centering
		Ours
	\end{minipage}
        \vspace{-2mm}
	\caption{\footnotesize \textbf{More qualitative results of our method (green line) and the GPV-Pose (blue line) on the REAL275 dataset.} We choose two instances from each scene. The ground truth results are shown with white lines. The estimated rotations of symmetric objects (\eg bowl, bottle, and can) are considered correct if the symmetry axis is aligned. }
\label{fig:qualitative}
\end{figure*}

More qualitative results comparing our method with the GPV-pose are shown in Fig.\ref{fig:qualitative}.



\newpage
{\small
\bibliographystyle{ieee_fullname}
\bibliography{supbib}
}